\documentclass{article}

\usepackage[utf8]{inputenc} 
\usepackage[T1]{fontenc}    
\usepackage{booktabs}       
\usepackage{nicefrac}       
\usepackage{microtype}      
\usepackage{times}             

\usepackage[round]{natbib}
\usepackage{amssymb,amsmath,amsthm,bbm}
\usepackage[margin=1in]{geometry}
\usepackage{verbatim,float,url,dsfont}
\usepackage{graphicx,subfigure,psfrag}
\usepackage{algorithm,algorithmic}
\usepackage{mathtools,enumitem}
\usepackage[colorlinks,citecolor=blue]{hyperref}
\usepackage[utf8]{inputenc} 
\usepackage[T1]{fontenc}    
\usepackage{hyperref}       
\usepackage{url}            
\usepackage{booktabs}       
\usepackage{amsfonts}       
\usepackage{nicefrac}       
\usepackage{microtype}      

\usepackage{times}             
\usepackage{amsthm}

\usepackage{bbm}
\usepackage{verbatim,float,dsfont}
\usepackage{psfrag}
\usepackage{algorithm,algorithmic}
\usepackage{mathtools,enumitem}
\mathtoolsset{showonlyrefs}
\usepackage{tcolorbox}
\usepackage{breqn,lipsum}
\usepackage{thmtools,thm-restate}
\usepackage{graphicx,stackengine}
\usepackage{array}
\usepackage{caption}
\usepackage{boldline}
\usepackage{amssymb}
\usepackage{multirow}

\newtheorem{theorem}{Theorem}
\newtheorem{lemma}[theorem]{Lemma}
\newtheorem{corollary}[theorem]{Corollary}
\newtheorem{proposition}[theorem]{Proposition}
\newtheorem{remark}[theorem]{Remark}
\newtheorem{definition}[theorem]{Definition}

\newtheorem{example}[theorem]{Example}

\DeclarePairedDelimiter\ceil{\lceil}{\rceil}
\DeclarePairedDelimiter\floor{\lfloor}{\rfloor}

\def\shownotes{1}  
\ifnum\shownotes=1
\newcommand{\authnote}[2]{$\ll$\textsf{\footnotesize #1 notes: #2}$\gg$}
\else
\newcommand{\authnote}[2]{}
\fi

\newcommand{\yw}[1]{\textcolor{red}{\textbf{[yuxiang: #1]}}}

\usepackage{accents}
\newcommand{\ubar}[1]{\underaccent{\bar}{#1}}

\makeatletter
\newcommand*\rel@kern[1]{\kern#1\dimexpr\macc@kerna}
\newcommand*\widebar[1]{%
  \begingroup
  \def\mathaccent##1##2{%
    \rel@kern{0.8}%
    \overline{\rel@kern{-0.8}\macc@nucleus\rel@kern{0.2}}%
    \rel@kern{-0.2}%
  }%
  \macc@depth\@ne
  \let\math@bgroup\@empty \let\math@egroup\macc@set@skewchar
  \mathsurround\z@ \frozen@everymath{\mathgroup\macc@group\relax}%
  \macc@set@skewchar\relax
  \let\mathaccentV\macc@nested@a
  \macc@nested@a\relax111{#1}%
  \endgroup
}
\makeatother

\newcommand{\ADAVAW}{\texttt{Ada-VAW}}

\newcommand{\argmin}{\mathop{\mathrm{argmin}}}
\newcommand{\argmax}{\mathop{\mathrm{argmax}}}

\def\R{\mathbb{R}}
\def\E{\mathbb{E}}

\def\Var{\mathrm{Var}}

\def\cA{\mathcal{A}}

\def\cE{\mathcal{E}}
\def\cF{\mathcal{F}}

\def\cH{\mathcal{H}}

\def\cS{\mathcal{S}}

\def\TV{\mathrm{TV}}

\def\bs{\ensuremath\boldsymbol}

\newcommand{\ud}{\mathrm d}
\newcommand{\red}[1]{\textcolor{red}{#1}}

\newcommand\tstrut{\rule{0pt}{3.0ex}}
\newcommand\bstrut{\rule[-2.0ex]{0pt}{0pt}}

\title{Adaptive Online Estimation of Piecewise Polynomial Trends}  

\author{Dheeraj Baby \\dheeraj@ucsb.edu \and Yu-Xiang Wang \\yuxiangw@cs.ucsb.edu}

\date{UC Santa Barbara}

\begin{document}

\maketitle

\begin{abstract}
We consider the framework of non-stationary stochastic optimization \citep{besbes2015non} with squared error losses and noisy gradient feedback where the dynamic regret of an online learner against a time varying comparator sequence is studied. Motivated from the theory of non-parametric regression, we introduce a \emph{new variational constraint} that enforces the comparator sequence to belong to a discrete $k^{th}$ order Total Variation ball of radius $C_n$. This variational constraint models comparators that have piece-wise polynomial structure which has many relevant practical applications \citep{tibshirani2014adaptive}. By establishing connections to the theory of wavelet based non-parametric regression, we design a \emph{polynomial time} algorithm that achieves the nearly \emph{optimal dynamic regret} of $\tilde{O}(n^{\frac{1}{2k+3}}C_n^{\frac{2}{2k+3}})$. The proposed policy is \emph{adaptive to the unknown radius} $C_n$. Further, we show that the same policy is minimax optimal for several other non-parametric families of interest.
\end{abstract}

\section{Introduction}\label{sec:problem-setup}

In time series analysis, estimating and removing the trend are often the first steps taken to make the sequence ``stationary''. The non-parametric assumption that the underlying trend is a piecewise polynomial or a spline \citep{deboorsplines}, is one of the most popular choices, especially when we do not know where the ``change points'' are and how many of them are appropriate. The higher order Total Variation (see Assumption A3) of the trend can capture in some sense both the sparsity and intensity of changes in underlying dynamics. A non-parametric regression method that penalizes this quantity --- trend filtering \citep{tibshirani2014adaptive} --- enjoys a superior \emph{local adaptivity} over traditional methods such as the Hodrick-Prescott Filter \citep{hodrick1997postwar}. However, Trend Filtering is an \emph{offline} algorithm which limits its applicability for the inherently \emph{online} time series forecasting problem. In this paper, we are interested in designing an online forecasting strategy that can essentially match the performance of the offline methods for trend estimation, hence allowing us to apply time series models forecasting on-the-fly. In particular, our problem setup (see Figure \ref{fig:proto}) and algorithm are applicable to all \emph{online variants} of trend filtering problem such as predicting stock prices, server payloads, sales etc. 

Let's describe the notations that will be used throughout the paper. All vectors and matrices will be written in bold face letters. For a vector $\bs x \in \mathbb{R}^m$, $\bs x[i]$ or $\bs x_i$ denotes its value at the $i^{th}$ coordinate. $\bs x[a:b]$ or $\bs{x_{a:b}}$ is the vector $[\bs x[a], \ldots, \bs x[b]]$.  $\|\cdot\|_p$ denotes finite dimensional $L_p$ norms. $\|\bs x\|_0$ is the number of non-zero coordinates of a vector $\bs x$. $[n]$ represents the set $\{1,\ldots,n \}$. $\bs{D}^i \in \mathbb{R}^{(n-i) \times n}$ denotes the discrete difference operator of order $i$ defined as in \citep{tibshirani2014adaptive} and reproduced below.
\begin{gather}
    \bs{D}^1
    =
    \begin{bmatrix}
    -1 & 1 & 0 &\dots & 0 & 0\\
    0 & -1 & 1 & \ldots & 0 & 0\\
    \vdots \\
    0 & 0 & 0 & \ldots & -1 & 1
    \end{bmatrix} \in \mathbb{R}^{(n-1) \times n},
\end{gather}
and $\bs D^{i} =  \bs{\tilde D}^1 \cdot \bs{D}^{i-1} \: \forall i \ge 2$ where $\bs{\tilde D}^1$ is the $(n-i) \times (n-i+1)$ truncation of $\bs D^{1}$.

The theme of this paper builds on the non-parametric online forecasting model developed in \citep{arrows2019}. We consider a sequential $n$ step interaction process between an agent and an adversary as shown in Figure \ref{fig:proto}.

\begin{figure}[h!]
	\centering
	\fbox{
		\begin{minipage}{13 cm}
\begin{enumerate}
    \item Fix a time horizon $n$.
    \item Agent declares a forecasting strategy $\cS$
    \item Adversary chooses a sequence $\bs \theta_{1:n}$
    \item For $t = 1,\ldots, n$:
    \begin{enumerate}
        \item Agent outputs a prediction $\cS(t)$.
        \item Adversary reveals $y_t = \bs \theta_{1:n}[t] + \epsilon_t$
    \end{enumerate}
    \item After $n$ steps, agent suffers a cumulative loss $\sum_{i=1}^n \big(\cS(i)- \bs \theta_{1:n}[i]\big)^2$.
\end{enumerate}
		\end{minipage}
	}
	\caption{Interaction protocol}
	\label{fig:proto}
\end{figure}
A forecasting strategy $\cS$ is defined as an algorithm that outputs a prediction $\cS(t)$ at time $t$ only based on the information available after the completion of time $t-1$. Random variables $\epsilon_t$ for $t \in [n]$ are independent and subgaussian with parameter $\sigma^2$. This sequential game can be regarded as an online version of the non-parametric regression setup well studied in statistics community. 

In this paper, we consider the problem of forecasting sequences that obey $n^{k} \|D^{k+1}\bs\theta_{1:n}\|_1 \le C_n$, $k \ge 0$ and $\|\bs\theta_{1:n}\|_\infty \le B$. The constraint $n^{k} \|D^{k+1}\bs\theta_{1:n}\|_1 \le C_n$ has been widely used in the rich literature of non-parametric regression. For example, the offline problem of estimating sequences obeying such higher order difference constraint from noisy labels under squared error loss is studied in \citep{locadapt,donoho1998minimax,tibshirani2014adaptive,graphtf, sadhanala2016total,guntuboyina2018constrainedTF} to cite a few. We aim to design forecasters whose predictions are only based on past history and still perform as good as a batch estimator that sees the entire observations ahead of time.

\begin{figure}[htp]
  \centering
  \stackunder{\hspace*{-0.5cm}\includegraphics[width=0.4\textwidth,height=0.7\textheight,keepaspectratio]{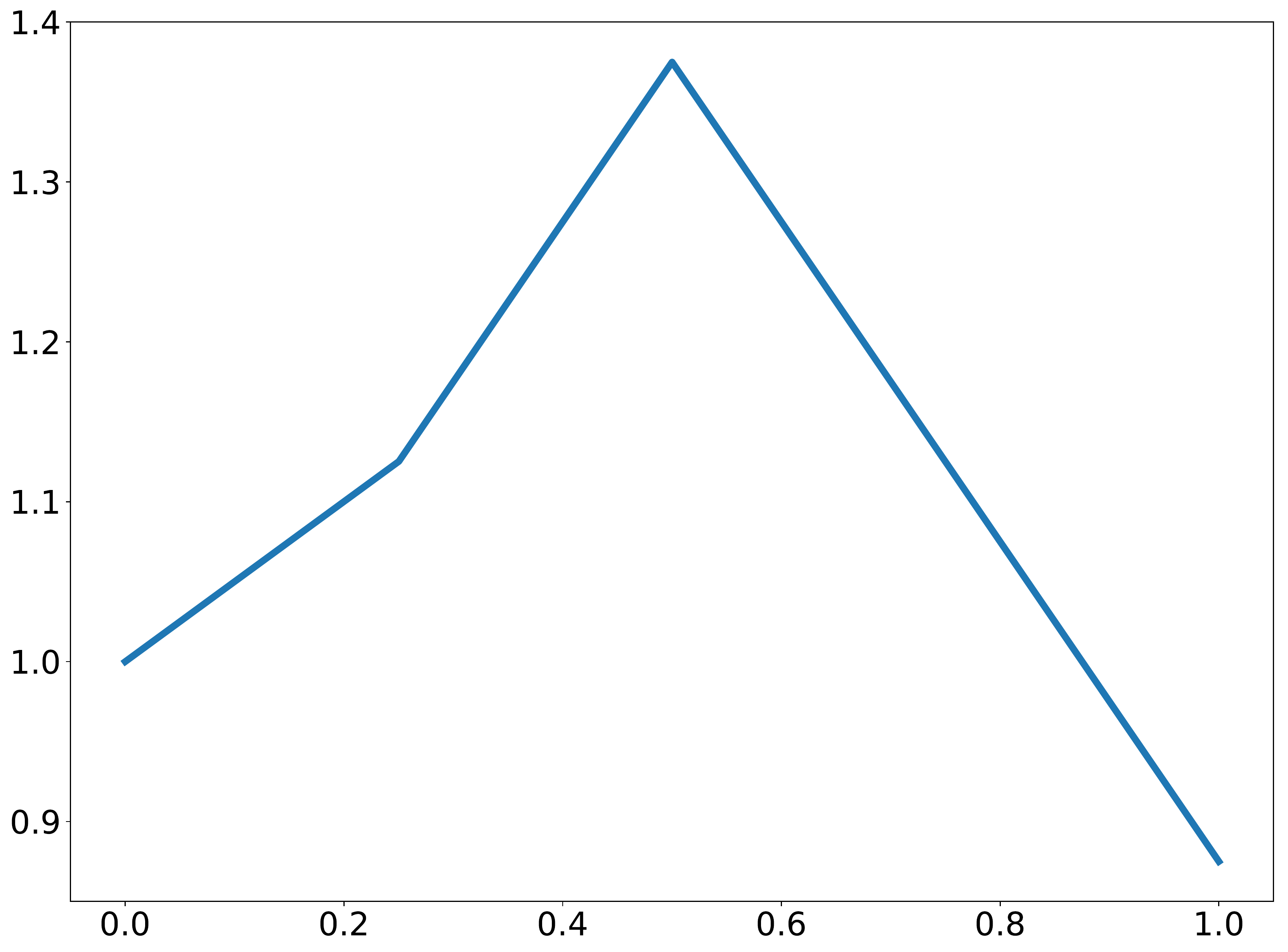}}{}%
  \stackunder{\includegraphics[width=0.6\textwidth,height=0.2\textheight,keepaspectratio]{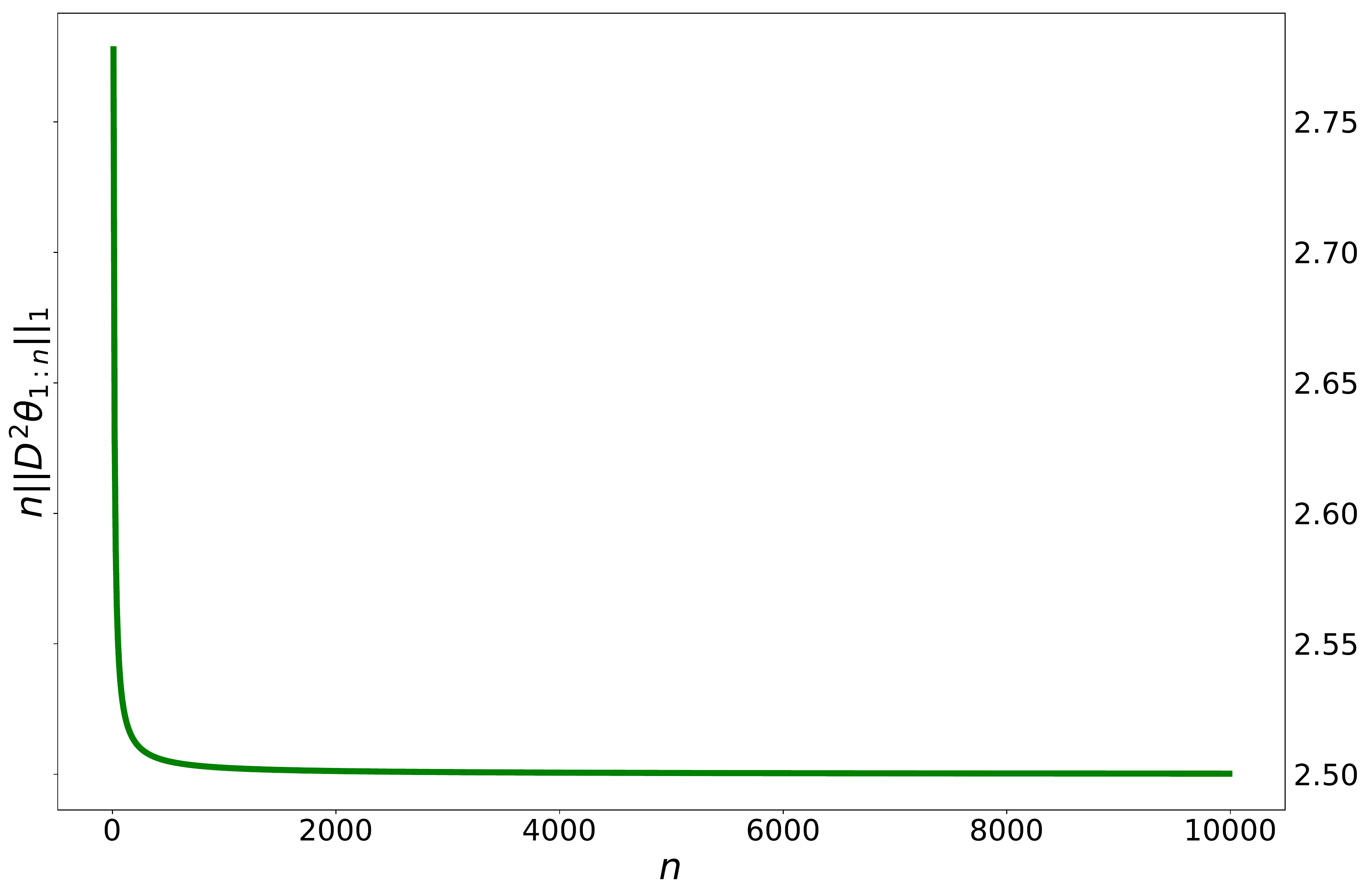}}{}
   \caption{\emph{A $TV^1$ bounded comparator sequence $\bs \theta_{1:n}$ can be obtained by sampling the continuous piecewise linear function on the left at points $i/n$, $i \in [n]$. On the right, we plot the $TV^1$ distance (which is equal to $n \|D^2 \bs \theta_{1:n} \|_1$ by definition) of the generated sequence for various sequence lengths $n$. As $n$ increases the discrete $TV^1$ distance converges to a constant value given by the continous $TV^1$ distance of the function on left panel. }}\label{fig:illustration}  
\end{figure}

\noindent\textbf{Scaling of $\bs{n^k}$.} 
The family $\{\bs\theta_{1:n} \;|\; n^{k} \|D^{k+1}\bs\theta_{1:n}\|_1 \le C_n\}$ may appear to be alarmingly restrictive for a constant $C_n$ due to the scaling factor $n^k$, but let us argue why this is actually a natural construct. The continuous $TV^k$ distance of a function $f:[0,1] \rightarrow \mathbb{R}$ is defined as $\int_{0}^{1} |f^{(k+1)}(x)|dx$, where $f^{(k+1)}$ is the $(k+1)^{th}$ order (weak) derivative. A sequence can be obtained by sampling the function at $x_i = i/n$, $i \in [n]$. Discretizing the integral yields the $TV^k$ distance of this sequence to be $n^{k}\|D^{k+1} \bs \theta_{1:n}\|_1$. Thus, the $n^{k}\|D^{k+1} \bs \theta_{1:n} \|_1$ term can be interpreted as the discrete approximation to continuous higher order TV distance of a function. See Figure \ref{fig:illustration} for an illustration for the case $k=1$.

\noindent\textbf{Non-stationary Stochastic Optimization.}
The setting above can also be viewed under the framework of non-stationary stochastic optimization as studied in \citep{besbes2015non,chen2018non} with squared error loss and noisy gradient feedback. At each time step, the adversary chooses a loss function $f_t(x) = (x - \bs \theta_t)^2$. Since $\nabla f_t(x) = 2 (x - \bs \theta_t)$, the feedback $\tilde \nabla f_t(x) = 2 (x - y_t)$ constitutes an unbiased estimate of the gradient $\nabla f_t(x)$. \citep{besbes2015non,chen2018non} quantifies the performance of a forecasting strategy $\cS$ in terms of dynamic regret as follows.
\begin{align}
    R_{dynamic}(\cS,\bs \theta_{1:n}) 
    &:=\E \left[ \sum_{t=1}^n f_t\left(\cS(t)\right) \right] - 
	\sum_{t=1}^n \inf_{x_t }f_t(x_t),
	= \E \left[\sum_{t=1}^n \left(\cS(t)- \bs \theta_{1:n}[t]\right)^2 \right], \label{eqn:dyn-regret}
\end{align}
where the last equality follows from the fact that when $f_t(x) = (x - \bs \theta_{1:n}[t])^2$, $\inf_x (x-\bs \theta_{1:n}[t])^2 = 0$. The expectation above is taken over the randomness in the noisy gradient feedback and that of the agent's forecasting strategy. It is impossible to achieve sublinear dynamic regret against arbitrary ground truth sequences. However if the sequence of minimizers  of  loss functions $f_t(x) = (x-\theta_t)^2$ obey a path variational constraint, then we can parameterize the dynamic regret as a function of the path length, which could be sublinear when the path-length is sublinear. Typical variational constraints considered in the existing work includes $\sum_t| \bs \theta_t- \bs \theta_{t-1}|$, $\sum_t|\bs \theta_t- \bs \theta_{t-1}|^2$, $(\sum_t\|f_t-f_{t-1}\|_p^q)^{1/q}$ \citep[see][for a review]{arrows2019}. These are all useful in their respective contexts, but do not capture higher order smoothness.

The purpose of this work is to connect ideas from batch non-parametric regression to the framework of online stochastic optimization and define a \emph{natural family of higher order variational functionals} of the form $\|D^{k+1} \bs \theta_{1:n}\|_1$ to track a comparator sequence with piecewise polynomial structure.
To the best of our knowledge such higher order path variationals for $k \ge 1$ are vastly unexplored in the domain of non-stationary stochastic optimization. In this work, we take the first steps in introducing such variational constraints to online non-stationary stochastic optimization and exploiting them to get sub-linear dynamic regret.





\section{Summary of results} \label{sec:summary}
In this section, we summarize the assumptions and main results of the paper.

\noindent\textbf{Assumptions.}
We start by listing the assumptions made and provide justifications for them.
\begin{enumerate}
\item[(A1)] The time horizon is known to be $n$.
\item[(A2)]  The parameter $\sigma^2$ of subgaussian noise in the observations is a known fixed positive constant.
\item[(A3)] The ground truth denoted by $\bs \theta_{1:n}$ has its $k^{th}$ order total variation bounded by some positive $C_n$, i.e., we consider ground truth sequences that belongs to the class 
$$\TV^{k}(C_n) := \{ \bs\theta_{1:n}\in \R^n :   n^{k}\|D^{k+1} \bs \theta_{1:n}\|_1 \leq C_n \}$$
We refer to $n^k \| D^{k+1} \bs \theta_{1:n}\|_1$ as $TV^k$ distance of the sequence $\bs \theta_{1:n}$. To avoid trivial cases, we assume $C_n = \Omega(1)$.
\item[(A4)] The TV order $k$ is a known \emph{fixed} positive constant.
\item[(A5)] $\|\bs \theta_{1:n}\|_{\infty} \le B$ for a known \emph{fixed} positive constant $B$.
\end{enumerate}

Though we require the time horizon to be known in advance in assumption (A1), this can be easily lifted using standard doubling trick arguments. The knowledge of time horizon helps us to present the policy in a most transparent way. If standard deviation of sub-gaussian noise is unknown, contrary to assumption (A2), then it can be robustly estimated by a Median Absolute Deviation estimator using first few observations, see for eg. \citet{DJBook}. This is indeed facilitated by the sparsity of wavelet coefficients of $TV^k$ bounded sequences. Assumption (A3) characterizes the ground truth sequences whose forecasting is the main theme of this paper. The $\TV^k(C_n)$ class features a rich family of sequences that can potentially exhibit spatially non-homogeneous smoothness. For example it can capture sequences that are piecewise polynomials of degree at most $k$. This poses a challenge to design forecasters that are \emph{locally adaptive} and can efficiently detect and make predictions under the presence of the non-homogeneous trends. Though knowledge of the TV order $k$ is required in assumption (A4), most of the practical interest is often limited to the lower orders $k=0,1,2,3$, see for eg. \citep{l1tf,tibshirani2014adaptive} and we present (in Appendix \ref{app:adapt}) a meta-policy based on exponential weighted averages \citep{BianchiBook2006} to adapt to these lower orders. Finally assumption (A5) is standard in the online learning literature.

\noindent\textbf{Our contributions.} We summarize our main results below.
\begin{itemize}
    \item When the revealed labels are noisy realizations of sequences that belong to $TV^k(C_n)$ we propose a \emph{polynomial time} policy called \ADAVAW{} (\textbf{Ada}ptive \textbf{V}ovk \textbf{A}zoury \textbf{W}armuth forecaster) that achieves the nearly \emph{minimax optimal} rate of $\tilde{O}\left(n^{\frac{1}{2k+3}}C_n^{\frac{2}{2k+3}}\right)$ for $R_{dynamic}$ with high probability. The proposed policy \emph{optimally adapts to the unknown radius} $C_n$.

    \item We show that the proposed policy achieves optimal $R_{dynamic}$ when revealed labels are noisy realizations of sequences residing in higher order discrete Holder and discrete Sobolev classes.
    
    \item When the revealed labels are noisy realizations of sequences that obey $\|D^{k} \bs \theta_{1:n}\|_0 \le J_n, \|\bs \theta_{1:n}\|_\infty \le B$, we show that the same policy achieves the minimax optimal $\tilde{O}(J_n)$ rate for for $R_{dynamic}$ with high probability. The policy  \emph{optimally adapts to unknown} $J_n$.

\end{itemize}

\noindent\textbf{Notes on key novelties.}
It is known that the VAW forecaster is an optimal algorithm for online polynomial regression with squared error losses \citep{BianchiBook2006}. With the side information of change points where the underlying ground truth switches from one polynomial to another, we can run a VAW forecaster on each of the stable polynomial sections to  control the cumulative squared error of the policy.  We use the machinery of wavelets to mimic an oracle that can provide side information of the change points. For detecting change points, a restart rule is formulated by exploiting connections between wavelet coefficients and locally adaptive regression splines. This is a \emph{more general} strategy than that used in \citep{arrows2019}. To the best of our knowledge, this is the \emph{first} time an interplay between VAW forecaster and theory of wavelets along with its adaptive minimaxity \citep{donoho1998minimax} has been used in the literature. 

Wavelet computations require the length of underlying data whose wavelet transform needs to be computed has to be a power of 2. In practice this is achieved by a padding strategy in cases where original data length is not a power of 2. We show that most commonly used padding strategies -- eg. zero padding as in \citep{arrows2019} -- are not useful for the current problem and propose a novel \emph{packing strategy} that alleviates the need to pad. This will be useful to many applications that use wavelets which can be well beyond the scope of the current paper. 

Our proof techniques for bounding regret use properties of the CDJV wavelet construction \citep{cdjv}. To the best of our knowledge, this is the \emph{first} time we witness the ideas from a general CDJV construction scheme implying useful results in an online learning paradigm. Optimally controlling the bias of VAW demands to carefully bound the $\ell_2$ norm of coefficients computed by polynomial regression. This is done by using ideas from number theory and symbolic determinant evaluation of polynomial matrices. This could be of independent interest in both offline and online polynomial regression.

\section{Related Work}
In this section, we briefly discuss the related work. A discussion on preliminaries and a detailed exposition of related literature is deferred to Appendix \ref{app:prelime} and \ref{app:lit} respectively. Throughout this paper when we refer as $\tilde{O}(n^{\frac{1}{2k+3}})$ as optimal regret we assume that $C_n = n^k\|D^{k+1} \bs \theta_{1:n}\|_1$ is $O(1)$.

\textbf{Non-parametric Regression} As noted in Section \ref{sec:problem-setup}, the problem setup we consider can be regarded as an online version of the batch non-parametric regression framework. It has been established (see for eg, \citep{locadapt,donoho1998minimax,tibshirani2014adaptive} that minimax rate for estimating sequences with bounded $TV^k$ distance under squared error loss scales as $n^{\frac{1}{2k+3}}(n^k\|D^{k+1} \bs \theta_{1:n}\|_1)^{\frac{2}{2k+3}}$ modulo logarithmic factors of $n$. In this work, we aim to achieve the same rate for minimax dynamic regret in online setting.

\textbf{Non-stationary Stochastic Optimization} Our forecasting framework can be considered as a special case of non-stationary stochastic optimization setting studied in \citep{besbes2015non,chen2018non}. It can be shown that their proposed algorithm namely, restarting Online Gradient Descend (OGD) yields a suboptimal dynamic regret of $O\left(n^{1/2}(\|D \bs \theta_{1:n}\|_1)^{1/2}\right)$ for our problem. However, it should be noted that their algorithm works with general strongly convex and convex losses. A summary of dynamic regret of various algorithms are presented in Table \ref{tab:TV-rates}. The rationale behind how to translate existing regret bounds to our setting is elaborated in Appendix \ref{app:lit}.

\begin{table*}[ht!]
\centering
\caption{\emph{Regret bounds for sequences that satisfy $n^k \|D^{k+1} \bs \theta_{1:n}\|_1 \le C_n$ with $\bs \theta[1:k+1] = 0$, $\|\bs \theta_{1:n}\|_\infty \le B$ and $k \ge 1$. The proposed policy doesn't require the knowledge of $C_n$ apriori while still attains the optimal dynamic regret modulo log factors. The bound for \ADAVAW{} holds even without the constraint on initial sequence values.}}\label{tab:TV-rates}
\resizebox{\columnwidth}{!}{
\begin{tabular}{|c|c|c|c|}
\hline
\textbf{Policy}                                              & \textbf{Dynamic Regret}              & \textbf{Known $C_n$?} & \textbf{Lower bound}                                  \\ \hline
\begin{tabular}[c]{@{}c@{}c@{}c@{}}Moving Averages,\\ Restarting OGD\\ \citep{besbes2015non}
\end{tabular} & $\tilde O(\sqrt{nC_n})$             & Yes                  & \multirow{5}{*}{ \begin{tabular}{@{}c@{}c@{}c@{}} \\ \\ \\ $\Omega\left(n^{1/(2k+3)}C_n^{2/(2k+3)}\right)$ \end{tabular}} \\ \cline{1-3}
\begin{tabular}[c]{@{}c@{}}OGD \\ \citep{zinkevich2003online}  \end{tabular}                                                         & $\tilde O(\sqrt{n C_n})$                & Yes                  &                                                       \\ \cline{1-3}
\begin{tabular}[c]{@{}c@{}} Ader \\ \citep{zhang2018adaptive}    \end{tabular}                                                    & $\tilde O(\sqrt{n C_n})$               & No                   &                                                       \\ \cline{1-3}
\begin{tabular}[c]{@{}c@{}} Arrows \\ \citep{arrows2019}     \end{tabular}                                                 & $\tilde{O}\left(n^{1/3}C_n^{2/3}\right)$           & No                   &                                                       \\ \cline{1-3}
\ADAVAW{} (\textbf{This paper})                                              & $\tilde{O}\left(n^{1/(2k+3)}C_n^{2/(2k+3)}\right)$ & No                   &                                                       \\ \hline
\end{tabular}
 }
\end{table*}

\textbf{Prediction of Bounded Variation sequences} Our problem setup is identical to that of \citep{arrows2019} except for the fact that they consider forecasting sequences whose zeroth order Total Variation is bounded. Our work can be considered as a generalization to any TV order $k$. Their algorithm gives a suboptimal regret of $O(n^{1/3}\|D\bs \theta_{1:n} \|_1^{2/3})$ for $k \ge 1$.

\textbf{Competitive Online Non-parametric Regression} \citep{rakhlin2014online} considers an online learning framework with squared error losses where the learner competes against the best function in a non-parametric function class. Their results imply via a non-constructive argument, the existence of an algorithm that achieves the regret of $\tilde{O}(n^{\frac{1}{2k+3}})$ for our problem.

\section{Main results}
We present below the main results of the paper. All proofs are deferred to the  appendix.
\subsection{Limitations of linear forecasters}
We exhibit a lower-bound on the dynamic regret that is implied by \citep{donoho1998minimax} in batch regression setting.

\begin{restatable}[Minimax Regret]{proposition}{propLower}
\label{prop:lower}
Let $y_t = \bs \theta_{1:n}[t] + \epsilon_t$ for $t=1,\ldots,n$ where $\theta_{1:n} \in TV^{(k)}(C_n)$, $|\bs \theta_{1:n}[t]| \le B$ and $\epsilon_t$ are iid $\sigma^2$ subgaussian random variables. Let $\cA_F$ be the class of all forecasting strategies whose prediction at time $t$ only depends on $y_1,\ldots,y_{t-1}$. Let $\bs{s_t}$ denote the prediction at time $t$ for a strategy $\bs s \in \cA_F$. Then,
\begin{align}
    \inf_{\bs s \in \cA_F} \sup_{\bs \theta_{1:n} \in TV^{(k)}(C_n)} \sum_{t=1}^{n} E\left[(\bs{s_t} - \bs \theta_{1:n}[t])^2 \right]
    = \Omega\left(\min\{n, n^{\frac{1}{2k+3}}C_n^{\frac{2}{2k+3}}\}\right),
\end{align}
where the expectation is taken wrt to randomness in the strategy of the player and $\epsilon_t$.
\end{restatable}

We define linear forecasters to be strategies that predict a fixed linear function of the history. This includes a large family of polices including the ARIMA family, Exponential Smoothers for Time Series forecasting, Restarting OGD etc. However in the presence of spatially inhomogeneous smoothness -- which is the case with TV bounded sequences -- these policies are doomed to perform sub-optimally. This can be made precise by providing a lower-bound on the minimax regret for linear forecasters. Since the offline problem of smoothing is easier than that of forecasting, a lower-bound on the minimax MSE of linear smoother will directly imply a lower-bound on the regret of linear forecasting strategies. By the results of \citep{donoho1998minimax}, we have the following proposition:

\begin{proposition}[Minimax regret for linear forecasters]
Linear forecasters will suffer a dynamic regret of at least $\Omega(n^{1/(2k+2)})$ for forecasting sequences that belong to $TV^k(1)$.
\end{proposition}

Thus we must look in the space of policies that are \emph{non-linear} functions of past labels to achieve a minimax dynamic regret that can potentially match the lower-bound in Proposition \ref{prop:lower}.

\subsection{Policy}
In this section, we present our policy and capture the intuition behind its design. First, we introduce the following notations.
\begin{itemize}
    \item The policy works by partitioning the time horizon into several bins. $t_h$ denotes start time of the current bin and $t$ be the current time point.
    \item $\bs W$ denotes the orthonormal Discrete Wavelet Transform (DWT) matrix obtained from a CDJV wavelet construction \citep{cdjv} using wavelets of regularity $k+1$.
    \item $T({\boldsymbol{y}})$ denotes the vector obtained by elementwise soft-thresholding of $\boldsymbol{y}$ at level $\sigma\sqrt{\beta \log l}$ where $l$ is the length of input vector.
    \item $\boldsymbol{x_t} \in \mathbb{R}^{(k+1)}$ denotes the vector $[1,t-t_h+k+1,\ldots,(t-t_h+k+1)^k]^T$.

    \item $A_t = \boldsymbol{I} + \sum_{s=t_h-k}^{t} \boldsymbol{x_s}\boldsymbol{x_s}^T$
    \item \texttt{recenter}$(\boldsymbol{y}[s:e])$ function first computes the Ordinary Least Square (OLS) polynomial fit with features $\boldsymbol{x}_s,\ldots,\boldsymbol{x}_e$. It then outputs the residual vector obtained by subtracting the best polynomial fit from the input vector $\boldsymbol{y}[s:e]$.
    \item Let $L$ be the length of a vector $\bs{u}_{1:t}$. \texttt{pack}($\bs{u}$) first computes $l = \floor{\log_2 L}$. It then returns the pair\\ $(\bs{u}_{1:2^l},\bs{u}_{t-2^l+1:t})$. We call elements of this pair as segments of $\bs{u}$.

\end{itemize}

\begin{figure}[h!]
	\centering
	\fbox{

		\begin{minipage}{13 cm}
\ADAVAW{}: inputs - observed $y$ values, TV order $k$, time horizon $n$, sub-gaussian parameter $\sigma$, range of ground truth $B$, hyper-parameter $\beta > 24$ and $\delta \in (0,1]$
\begin{enumerate}
    \item For $ t = 1$ to $k-1$, predict 0
    \item Initialize $t_h = k$
    \item For $t$ = $k$ to $n$:
    \begin{enumerate}
        \item Predict $\hat{y_t} = \langle \boldsymbol{x_t},A_t^{-1}\sum_{s=t_h-k}^{t-1}y_s \boldsymbol{x_s} \rangle$
        \item Observe $y_t$ and suffer loss $(\hat{y}_t - \bs \theta_{1:n}[t])^2$
        \item Let $\mathbf{y}_r = $\texttt{recenter}$(\mathbf{y}[t_h-k:t])$ and $L$ be its length
        \item Let $(y_1,y_2) = \texttt{pack}(\bs{y}_r)$
        \item Let $(\hat{\boldsymbol{\alpha}}_1,\hat{\boldsymbol{\alpha}}_2) = (T(\bs  W\mathbf{y}_1),T(\bs  W\mathbf{y}_2))$
        \item Restart Rule: If $\|\hat{\bs{\alpha}}_1\|_2 + \|\hat{\bs{\alpha}}_2\|_2 > \sigma$ then   \begin{enumerate}
            \item set $t_h = t+1$
        \end{enumerate}
    \end{enumerate}
\end{enumerate}
		\end{minipage}
	}
\end{figure}

The basic idea behind the policy is to adaptively detect intervals that have low $TV^k$ distance. If the $TV^k$ distance within an interval is guaranteed to be low enough, then outputting a polynomial fit can suffice to obtain low prediction errors. Here we use the polynomial fit from  VAW \citep{vovk2001} forecaster in step 3(a) to make predictions in such low $TV^k$ intervals. Step 3(e) computes denoised wavelets coefficients. It can be shown that the expression on the LHS of the inequality in step 3(f) can be used to lower bound  $\sqrt{L}$ times the $TV^k$ distance of the underlying  ground truth with high probability. Informally speaking, this is expected as the wavelet coefficents for a CDJV system with regularity $k$ are computed using higher order differences of the underlying signal. A restart is triggered when the scaled $TV^k$ lower-bound within a bin exceeds the threshold of $\sigma$. Thus we use the energy of denoised wavelet coefficients as a device to detect low $TV^k$ intervals. In Appendix \ref{app:pad_problems} we show that popular padding strategies such as zero padding, greatly inflate the $TV^k$ distance of the  recentered sequence for $k \ge 1$. This hurts the dynamic regret of our policy. To obviate the necessity to pad for performing the DWT, we employ a packing strategy as described in the policy.

\subsection{Performance Guarantees}
\begin{restatable}{theorem}{thmMain}
\label{thm:main}

Consider the the feedback model $y_t = \bs \theta_{1:n}[t] + \epsilon_t$ $t=1,\ldots,n$ where $\epsilon_t$ are independent $\sigma^2$ subguassian noise and $|\bs \theta_{1:n}[t]| \le B$. If $\beta = 24+\frac{8\log(8/\delta)}{\log(n)}$, then with probability at least $1-\delta$, \ADAVAW{} achieves a dynamic regret of $\tilde{O}\left(n^{\frac{1}{2k+3}}\left(n^k \|D^{k+1} \bs \theta_{1:n}\|_1 \right)^{\frac{2}{2k+3}}\right)$ where $\tilde{O}$ hides poly-logarithmic factors of $n$, $1/\delta$ and constants $k$,$\sigma$,$B$ that do not depend on $n$.
\end{restatable}
\begin{proof}[Proof Sketch]
Our proof strategy falls through the following steps.
\begin{enumerate}
    \item Obtain a high probability bound of bias variance decomposition type on the total squared error incurred by the policy within a bin.
    \item Bound the variance by optimally bounding the number of bins spawned.
    \item Bound the squared bias using the restart criterion.
\end{enumerate}

Step 1 is achieved by using the subgaussian behaviour of revealed labels $y_t$. For step 2, we first connect the wavelet coefficients of a recentered signal to its $TV^k$ distance using ideas from theory of Regression Splines. Then we invoke the ``uniform shrinkage'' property of soft thresholding estimator to construct a lowerbound of the $TV^k$ distance within a bin. Such a lowerbound when summed across all bins leads to an upperbound on the number of bins spawned. Finally for step 3, we use a reduction from the squared bias within a bin to the regret of VAW forecaster and exploit the restart criterion and adpative minimaxity of soft thresholding estimator \citep{donoho1998minimax} that uses a CDJV wavelet system.
\end{proof}

\begin{corollary}
Consider the setup of Theorem \ref{thm:main}. For the problem of forecasting sequences $\bs \theta_{1:n}$ with $n^k \|D^{k+1} \bs \theta_{1:n}\|_1 \le C_n$ and $\|\bs \theta_{1:n}\|_\infty \le B$, \ADAVAW{} when run with $\beta = 24+\frac{8\log(8/\delta)}{\log(n)}$ yields a dynamic regret of $\tilde{O}\left(n^{\frac{1}{2k+3}}\left(C_n\right)^{\frac{2}{2k+3}}\right)$ with probability atleast $1-\delta$.
\end{corollary}

\begin{remark} (\emph{Adaptive Optimality}) 
By combining with trivial regret bound of $O(n)$, we see that dynamic regret of \ADAVAW{} matches the lower-bound provided in Proposition \ref{prop:lower}. \ADAVAW{} optimally adapts to the variational budget $C_n$. Adaptivity to time horizon $n$ can be achieved by the standard doubling trick.
\end{remark}

\begin{remark}(\emph{Extension to higher dimensions})\label{rem:highdim}
Let the ground truth $\bs \theta_{1:n}[t] \in \mathbb{R}^d$ and let $ \bs v_i = [\bs \theta_{1:n}[1][i],\ldots,\bs \theta_{1:n}[n][i]], \Delta_i = n^k \|D^{k+1}\bs v_i\|_1$ for each $i \in [d]$. Let $\sum_{i=1}^{d} \Delta_{i} \le C_n$. Then by running $d$ instances of \ADAVAW{} in parallel where instance $i$ predicts ground truth sequence along co-ordinate $i$, a regret bound of $\tilde{O} \left(d^{\frac{2k+1}{2k+3}} n^{\frac{1}{2k+3}} C_n^{\frac{2}{2k+3}} \right)$ can be achieved.
\end{remark}

\begin{remark}(\emph{Generalization to other losses}) \label{rem:genloss}
Consider the protocol in Figure \ref{fig:proto}. Instead of squared error losses in step (5), suppose we use loss functions $f_t(x)$ such that $\argmin f_t(x) = \bs \theta_{1:n}[t]$ and $f_t'(x)$ is $\gamma$-Lipschitz. Under this setting, \ADAVAW{} yields a dynamic regret of $\tilde{O}\left(\gamma n^{\frac{1}{2k+3}} C_n^{\frac{2}{2k+3}}\right)$ with probability at least $1-\delta$. Concrete examples include (but not limited to):
\begin{enumerate}
    \item Huber loss, $f_t^{(\omega)}(x) = \begin{cases} 
      0.5(x - \bs \theta_{[1:n]}[t])^2 & |x - \bs \theta_{[1:n]}[t]| \le \omega \\
      \omega(|x - \bs \theta_{[1:n]}[t]| - \omega/2) & \text{ otherwise}
   \end{cases}$ is 1-Lipschitz in gradient.
   \item Log-Cosh loss,  $f_t(x) = \log(\cosh(x - \bs \theta_{[1:n]}[t]))$ is 1-Lipschitz in gradient.
   \item $\epsilon$-insensitive logistic loss \citep{dekel2005loss}, $f_t^{(\epsilon)}(x) = \log(1+e^{x - \bs \theta_{[1:n]}[t] - \epsilon}) + \log(1+e^{-x + \bs \theta_{[1:n]}[t] - \epsilon}) - 2 \log (1 + e^{-\epsilon})$ is 1/2-Lipschitz in gradient.
\end{enumerate}
\end{remark}

The rationale behind both Remark \ref{rem:highdim} and Remark \ref{rem:genloss} is described at the end of Appendix \ref{app:regret}

\begin{restatable}{proposition}{propRuntime}
\label{prop:runtime}
There exist an $O\left(((k+1) n)^2\right)$ run-time implementation of \ADAVAW{}.
\end{restatable}

The run-time of $O(n^2)$ is larger than the $O(n\log n)$ run-time of the more specialized algorithm of \citep{arrows2019} for $k=0$. This is due to the more complex structure of higher order CDJV wavelets which invalidates their trick that updates the Haar wavelets in an amortized $O(1)$ time. 

\section{Extensions}
In this section, we discuss the potential applications of the proposed algorithm which broadens its generalizability to several interesting use cases.

\subsection{Optimality for  Higher Order Sobolev and Holder Classes}

So far we have been dealing with total variation classes, which can be thought of as $\ell_1$-norm of the $(k+1)$th order derivatives. An interesting question to ask is ``how does \ADAVAW{} behave under smoothness metric defined in other norms, e.g., $\ell_2$-norm and $\ell_{\infty}$-norm?''
Following \citep{tibshirani2014adaptive}, we define the higher order discrete Sobolev class $\cS^{k+1}(C_n')$ and discrete Holder class $\cH^{k+1}(L_n')$ as follows.
\begin{align}
    \cS^{k+1}(C_n')
    &= \{\bs \theta_{1:n} : n^{k}\|D^{k+1}\bs \theta_{1:n}\|_2 \le C_n'\},\\
    \cH^{k+1}(L_n')
    &= \{\bs \theta_{1:n} : n^{k}\|D^{k+1}\bs \theta_{1:n}\|_\infty \le L_n'\},
\end{align}
where $k \ge 0$. These classes feature sequences that are \emph{spatially more regular} in comparison to the higher order $TV^k$ class. It is well known that (see for eg. \citep{gyorfi2002book}) the following embedding holds true:
\begin{align}
  \cH^{k+1}\left(\frac{C_n}{n}\right)
    \subseteq  \cS^{k+1}\left(\frac{C_n}{\sqrt{n}}\right)
    &\subseteq TV^k(C_n).
\end{align}
Here $\frac{C_n}{\sqrt{n}}$  and $\frac{C_n}{n}$ are respectively the maximal radius of a Sobolev ball and Holder ball enclosed within a $TV^k(C_n)$ ball. Hence we have the following Corollary.
\begin{corollary}
Assume the observation model of Theorem \ref{thm:main} and that $\bs \theta_{1:n} \in \cS^{k+1}(C_n')$. If $\beta = 24+\frac{8\log(8/\delta)}{\log(n)}$, then with probability at least $1-\delta$, \ADAVAW{} achieves a dynamic regret of $\tilde{O}\left(n^{\frac{2}{2k+3}}[C_n']^{\frac{2}{2k+3}}\right)$.
\end{corollary}
It turns out that this is the optimal rate for the Sobolev classes, even in the easier, offline non-parametric regression setting \citep{gyorfi2002book}. Since a Holder class can be sandwiched between two Sobolev balls of same minimax rates \citep[see, e.g.,][]{gyorfi2002book}, this also implies the adaptive optimality for the Holder class. We emphasize that \ADAVAW{} does not need to know the $C_n, C_n'$ or $L_n'$ parameters, which implies that it will achieve the smallest error permitted by the right norm that captures the smoothness structure of the unknown sequence $\bs \theta_{1:n}$.


\subsection{Optimality for the case of Exact Sparsity} \label{sec:exact_sparse}

Next, we consider the performance of \ADAVAW{} on sequences satisfying an $\ell_0$-(pseudo)norm measure of the smoothness, defined as
\begin{align}
    \cE^{k+1}(J_n)
    &= \{\bs \theta_{1:n} : \|D^{k+1} \bs \theta_{1:n}\|_0 \le J_n, \|\bs \theta_{1:n}\|_\infty \le B\}.
\end{align}
This class captures sequences that has at most $J_n$ jumps in its $(k+1)^{th}$ order difference,
%
which covers (modulo the boundedness) $k$th order discrete splines \citep[see, e.g.,][Chapter 8.5]{schumaker2007spline} with exactly $J_n$ knots, and arbitrary piecewise polynomials with $O(J_n/k)$ polynomial pieces.

The techniques we developed in this paper allows us to establish the following performance guarantee for \ADAVAW{}, when applied to sequences in this family. 


\begin{restatable}{theorem}{thmExact}
\label{thm:exact-sparse}
Let $y_t = \bs \theta_{1:n}[t] + \epsilon_t$, for $t=1,\ldots,n$ where $\epsilon_t$ are iid sub-gaussian with parameter $\sigma^2$ and $\|D^{k+1}\bs \theta_{1:n}\|_0 \le J_n$ with $|\bs \theta_{1:n}[t]| \le B$ and $J_n \ge 1$. If $\beta = 24+\frac{8\log(8/\delta)}{\log(n)}$, then with probability at least $1-\delta$, \ADAVAW{} achieves a dynamic regret of $\tilde{O}\left(J_n\right)$ where $\tilde{O}$ hides polynmial factors of $\log(n)$ and $\log(1/\delta)$.
\end{restatable}
We also establish an information-theoretic lower bound that applies to all algorithms. 
\begin{restatable}{proposition}{propExactLower}
\label{prop:exact-sparse}
Under the interaction model in Figure \ref{fig:proto}, the minimax dynamic regret for forecasting sequences in $\cE^{k+1}(J_n)$ is $\Omega(J_n)$.
\end{restatable}

\begin{remark}
Theorem~\ref{thm:exact-sparse} and Proposition~\ref{prop:exact-sparse} imply that \ADAVAW{} is  optimal (up to logarithmic factors) for the sequence family $\cE^{k}(J_n)$. It is noteworthy that the \ADAVAW{} is adaptive in $J_n$, so it is essentially performing as well as an oracle that knows \emph{how many} knots are enough to represent the input sequence as a discrete spline and \emph{where} they are in advance (which leaves only the $J_n$ polynomials to be fitted). 
\end{remark}

\section{Conclusion}
In this paper, we considered the problem of forecasting $TV^k$ bounded sequences and proposed the first efficient algorithm -- \ADAVAW{}-- that is adaptively minimax optimal. We also discussed the adaptive optimality of  \ADAVAW{} in various parameters and other function classes.  In establishing strong connections between the locally adaptive nonparametric regression literature to the adaptive online learning literature in a concrete problem, this paper could serve as a stepping stone for future exchanges of ideas between the research communities, and hopefully spark new theory and practical algorithms.


\section*{Acknowledgment}
The research is partially supported by a start-up grant from UCSB CS department, NSF Award \#2029626 and generous gifts from Adobe and Amazon Web Services.

\section*{Broader Impact}

\begin{enumerate}
    \item Who may benefit from the research? This work can be applied to the task of estimating trends in time series forecasting. For example, financial firms can use it to do stock market predictions, distribution sector can use it do inventory planning, meterological observatories can use it for weather forecast and health and planning sector can forecast the spread of contagious diseases etc.
    \item Who may be put at disadvantage? Not applicable
    \item What are the consequences of failure of the system? There is no system to speak off, but failure of the strategy can lead to financial losses for the firms deploying the strategy to do forecasting. Under the assumptions stated in the paper though, the technical results are formally proven and come with the stated mathematical guarantee.
    \item Method leverages the biases in data? Not applicable.
\end{enumerate}

\bibliography{tf,yx}
\bibliographystyle{plainnat}

\newpage 

\appendix

\section{Background} \label{app:prelime}
In this section, we compile some preliminary results well established in literature. For brevity we only discuss the essential aspects that lead to design of our algorithm and its proof.
\subsection{Non-parametric regression}
A popular model studied in non-parametric regression is
\begin{align}
    y_i = f(i/n) + \epsilon_i, i \in [n], \label{eq:nonpar}
\end{align}
where $\epsilon_i$ are iid subgaussian noise and for unknown $f:[0,1] \rightarrow \mathbb{R}$. The idea is to recover the underlying ground truth $f$ from the observations $y_i$. Let $\bs \theta_{1:n} = [f(1/n),\ldots,f(1)] \in \mathbb{R}^n$ be the ground truth sequence. We constraint the ground truth to belong to some non-parametric class. A well studied (dating back since 90s atleast) non-parametric family is the class of $TV^k$ bounded sequences defined below.

\begin{align}
    \TV^{k}(C_n) := \{ \bs\theta_{1:n}\in \R^n :   n^{k}\|D^{k+1} \bs \theta_{1:n}\|_1 \leq C_n \}.
\end{align}

\begin{figure}[htp]
  \centering
  \stackunder{\hspace*{-0.5cm}\includegraphics[width=0.4\textwidth,height=0.7\textheight,keepaspectratio]{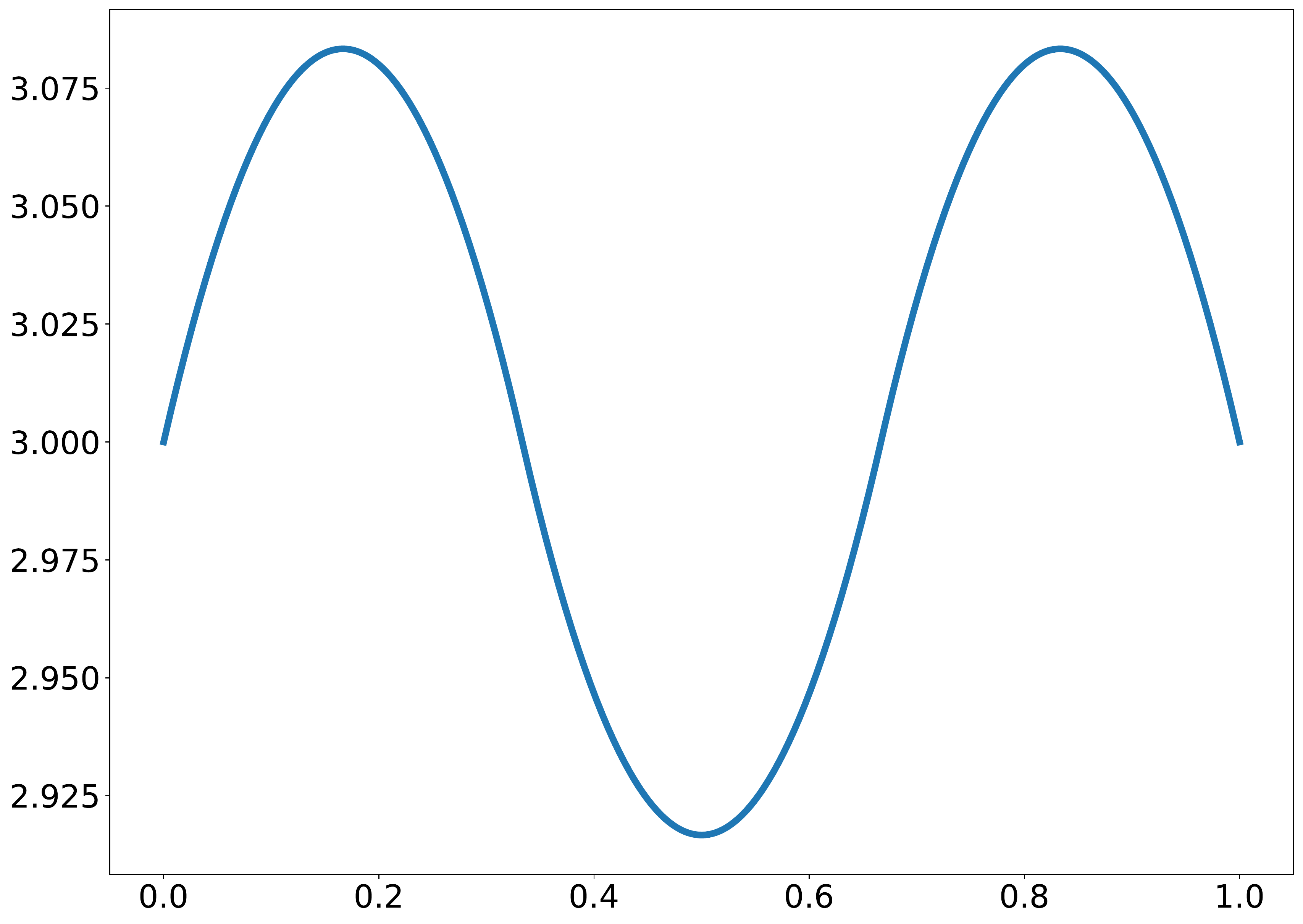}}{}%
  \stackunder{\includegraphics[width=0.6\textwidth,height=0.2\textheight,keepaspectratio]{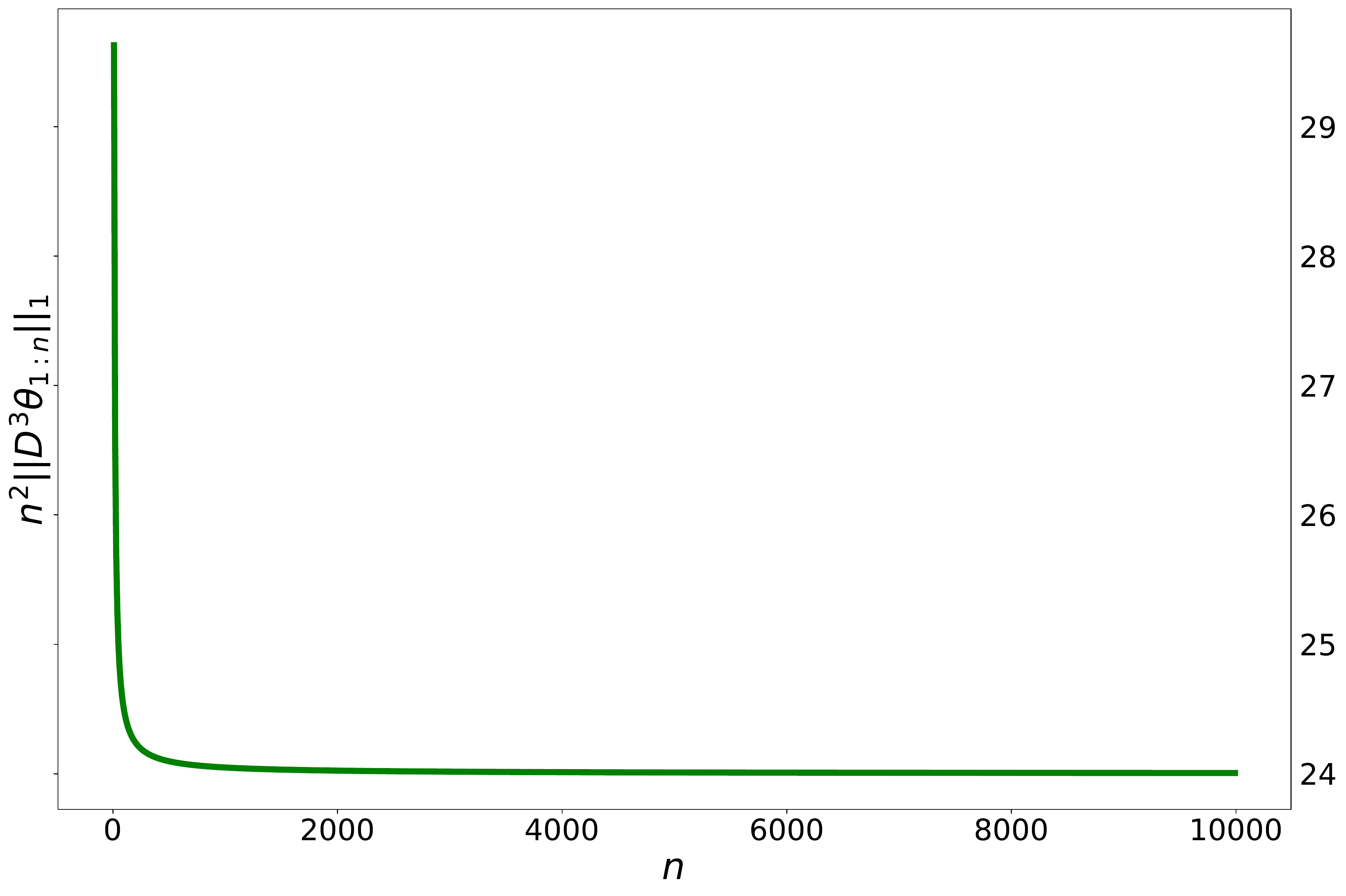}}{}
   \caption{\emph{A $TV^2$ bounded sequence $\bs \theta_{1:n}$ can be obtained by sampling the continuous piecewise quadratic function on the left at points $i/n$, $i \in [n]$. On the right, we plot the $TV^2$ distance of the generated sequence for various sequence lengths $n$. As $n$ increases the discrete $TV^2$ distance converges to a constant value given by the continous $TV^2$ distance of the function on left panel. }}\label{fig:illustration2}  
\end{figure}

The sequences in this class have a piecewise (discrete) polynomial structure. Each stable section features a polynomial of degree atmost $k$. However the number of polynomial sections and positions where the sequence transitions from one polynomial to another is unknown. This makes the task of estimating ground truth from noisy observations quite challenging. Moreover as noted in \citep{l1tf}, such sequences can be used to model a wide spectrum of real world phenomena. As noted in Section \ref{sec:summary}, such $TV^k$ sequences can be obtained by sampling the function whose continuous $TV^k$ distance is bounded. An illustration for $k=2$ is given in Figure \ref{fig:illustration2}.

The purpose of a non-parametric regression algorithm $\cA$ is to estimate $\bs \theta_{1:n}$ given the noisy observations $y_i$. The most common metric used to ascertain the performance of an algorithm in non-parametric regression literature is the squared error loss. Let the estimates of the algorithm be $\hat{\bs y}_{1:n}$. The empirical risk is defined as

\begin{align}
    R_n = E \left[\sum_{t=1}^{n} (\hat{\bs y}_{1:n}[t] - \bs \theta_{1:n}[t])^2\right],
\end{align}
and the minimax risk for estimating sequences in $TV^k(C_n)$ is formulated as
\begin{align}
    R_n^* = \min_{\cA} \max_{\bs \theta \in TV^{k}(C_n)} R_n,
\end{align}
where $\cA$ is an estimation of algorithm. It is well established (see eg. \citep{donoho1998minimax}) that 
\begin{align} \label{eq:minmaxlb}
    R_n^* = \Omega(n^{\frac{1}{2k+3}}C_n^{\frac{2}{2k+3}}).
\end{align}

\subsection{Wavelet Smoothing}
Let $\mathbb{Z}_+ = \mathbb{N} \cup \{ 0 \}$ and $L_2[0,1]$ be the space of all square integrable functions defined in $[0,1]$.

\begin{definition}
A Multi Resolution Analysis (MRA) on interval [0,1] is a sequence of subspaces $\{V_j, j \in \mathbb{Z}_+ \}$ satisfying
\begin{enumerate}
    \item $V_j \subset V_{j+1}$
    \item $f(x) \in V_j$ if and only if $f(2x) \in V_{j+1}$ 
    \item $\bigcap_{j \in \mathbb{Z}_+} = \{ 0 \}$ and $\bigcup_{j \in \mathbb{Z}_+}$ spans $L_2[0,1]$.
    \item There exists a function $\phi \in V_0$ such that $\{ \phi(x-k) : k \in \mathbb{Z} \text{ such that } \phi(x-k) \text{ is supported in [0,1]} \}$ is an orthonormal basis for $V_0$
\end{enumerate}
\end{definition}

The spaces $V_j$ form an increasing sequence of approximations to $L_2[0,1]$. Let $\phi_{jk}(x) = 2^{j/2}\phi(2^jx - k)$. In what follows we define $\phi_{jk}(x) = 0$ if it is not supported entirely within $[0,1]$. Due to properties 2 and 4 it follows that $\{ \phi_{jk}(x), k \in \mathbb{Z}\}$ is an orthonormal basis for $V_j$. The function $\phi(x)$ is called the \emph{scale function}.

Now let's define wavelets. \emph{Detail subpace} $W_j \subset L_2[0,1]$ is defined as the orthogonal complement of $V_j$ in $V_{j+1}$. A function $\psi(x)$ is defined to be a \emph{wavelet} (or mother wavelet) function if $\{ \psi_{jk}(x) = 2^{j/2}\psi(2^jx - k), k \in \mathbb{Z}_+ \text{ such that } \psi_{jk}(x) \text{ is supported in } [0,1] \}$ is an orthonormal basis for $W_j$ $\forall j \in \mathbb{Z}_+$.

\begin{definition}
A wavelet function $\psi(x)$ has regularity $r$ if
\begin{align}
    \int_{0}^{1} x^p \psi(x) dx = 0, p = 0,\ldots,r-1.
\end{align}
\end{definition}

The CDJV construction in \citep{cdjv} is an algorithm that provides a scale function $\phi(x)$ and wavelet function $\psi(x)$ of a given regularity $r$. We record an important property of this construction.

\begin{proposition} \label{prop:cdjv}
The CDJV construction with regularity $r$ satisfy
\begin{enumerate}
    \item Let $L = \ceil{\log 2r}$. Then $V_L$ contains polynomials of degree $\le r-1$.
    \item The functions $\psi_{jk}(x), j\ge L, k \in \mathbb{Z}$ are orthogonal to polynomials of degree atmost $r-1$.
\end{enumerate}
\end{proposition}

Let $n = 2^J$ and $L < J$. A discrete Wavelet Transform (DWT) matrix $\bs W \in \mathbb{R}^{n \times n}$ is generated by sampling the basis functions that make up $V_L$ and $W_L,\ldots,W_{J-1}$ at points $i/n, i \in [n]$ and scaling them by a factor of $n^{-1/2}$. The obtained matrix $\bs W$ can be shown to be orthonormal. The total number of basis functions that make up the space $V_J$ is $n$.

Now to provide a clearer picture, we orchestrate all the above ideas with the help of the simple \emph{Haar wavelets}.

\begin{definition}
The Haar MRA on [0,1] is defined by
\begin{enumerate}
    \item The scale function $\phi(x) = 1$
    \item The mother wavelet $\psi(x) = -1 \text{ if } x \le 1/2; 1$ otherwise.
    \item Both $\phi(x), \psi(x)$ are zero outside $[0,1]$
\end{enumerate}
\end{definition}

Here $V_0$ is the space of constant signals in $[0,1]$. $W_0$ is the functions of the form $c \psi(x)$ for $c \in \mathbb{R}$. $W_1$ is spanned by $\psi_{10}(x)$ and $\psi_{11}(x)$ and so on. It is clear that regularity of Haar wavelet $\psi(x)$ is 1. In fact Haar system is a special case of CDJV construction for regularity 1. Hence $L = \ceil{\log 2r} = 1$. The space $V_1$ is spanned by $\{\phi(x), \psi(x) \}$. It is easy to verify that space $V_1$ contains all polynomials of degree $r-1 = 0$ as asserted by Proposition \ref{prop:cdjv}. Furthermore property 2 stated in Proposition \ref{prop:cdjv} is also true.

Now let's construct the orthonormal Haar DWT matrix $\bs W \in \mathbb{R}^{n \times n}$. Let $J = \log n$ We need to sample sample basis functions of $V_1,W_1,\ldots W_{J-1}$ at points $i/n, i \in [n]$ and scale them by $n^{-1/2}$. For simplicity we illustrate this for $n=4$.

\begin{gather}
    \bs{W}
    =
    \begin{bmatrix}
    1/2 & 1/2 & 1/2 & 1/2\\
    1/2 & 1/2 & -1/2 & -1/2\\
    1/\sqrt{2} & -1/\sqrt{2} & 0 & 0\\
    0 & 0 & 1/\sqrt{2} & -1/\sqrt{2}
    \end{bmatrix}.
\end{gather}

It is noteworthy that general CDJV wavelets for regularity $r \ge 2$ do not have a closed form expression like the Haar system. The filter coefficients are computed by an efficient iterative algorithm.

Define the soft thresholding operator as
\begin{align}
    T_\lambda(x) 
    &= 
    \begin{cases}
    0 & |x| \le \lambda \\
    x - \lambda  & x > \lambda\\
    x + \lambda & x < \lambda
    \end{cases}
\end{align}

If the input is a vector the operation is done co-ordinate wise.

Now we are ready to discuss the famous universal soft thresholding algorithm of \citep{donoho1998minimax}.

\begin{figure}[h!]
	\centering
	\fbox{
		\begin{minipage}{13 cm}
		WaveletSoftThreshold: Inputs - observations $\bs y_{1:n}$, subgaussian parameter $\sigma$ of noise in \eqref{eq:nonpar}, TV order $k$
\begin{enumerate}
    \item Let $\bs W \in \mathbb{R}^{n \times n}$ be a CDJV DWT  matrix of regularity $k+1$. 
    \item Output $\hat{\bs y}_{1:n} = \bs{W}^T T_{\sigma\sqrt{2 \log n}}(\bs Wy)$.
\end{enumerate}
		\end{minipage}
	}
\end{figure}

We have the following proposition due to \citep{donoho1998minimax}.

\begin{proposition}
The risk of the wavelet soft thresholding scheme satisfy
\begin{align}
    R_n = \tilde{O}(n^{\frac{1}{2k+3}}C_n^{\frac{2}{2k+3}}).    
\end{align}
\end{proposition}

Comparing with equation \eqref{eq:minmaxlb} we see that WaveletSoftThreshold is a near minimax algorithm for estimating sequences in $TV^k(C_n)$. It optimally adapts to the unknown radius $C_n$ as well.

\subsection{Vovk Azoury Warmuth (VAW) forecaster}
The VAW algorithm is shown in Figure \ref{fig:vaw}. For a more elaborate discussion on this algorithm, refer to chapter 11 of \citep{BianchiBook2006}. The VAW forecaster is defined as follows.

\begin{figure}[h!]
	\centering
	\fbox{
		\begin{minipage}{13 cm}
		VAW algorithm
\begin{enumerate}
    \item Adversary reveals $\bs x_t \in \mathbb{R}^d$.
    \item Agent predicts $\hat{p}_t = \hat{\bs{w}}_{t-1}^T \bs x_t$ with $\hat{\bs {w}}_t = (\bs I + \sum_{s=1}^{t}\bs x_s \bs x_s^T)^{-1} \sum_{s=1}^{t-1}y_s \bs x_s$.
    \item Adversary reveals $y_t$.
    \item Incur loss $(\hat{p}_t - y_t)^2$.
\end{enumerate}
		\end{minipage}
	}
	\caption{The VAW algorithm}
	\label{fig:vaw}
\end{figure}

We have the following guarantee on the regret bound of VAW.
\begin{proposition}
If the VAW forecaster is run on a sequence $(\bs x_1, y_1), \ldots, (\bs x_n,y_n) \in \mathbb{R}^d \times \mathbb{R}$, then for all $\bs u \in \mathbb{R}^d$ and $n \ge 1$,
\begin{align}
\sum_{t=1}^{n} (y_t - \hat{p}_t)^2 - (y_t - \bs{u}^T\bs{x}_t)
&\le \frac{1}{2}\| \bs u\|_2^2 + \frac{dY^2}{2} \log(1+\frac{nX^2}{d}),
\end{align}   
where $\| \bs x_t \|_2 \le X$, and $|y_t| \le Y, t \in [n]$.
\end{proposition}

\section{Detailed Discussion of Related Literature}
\label{app:lit}
In this section, we discuss the connections of our work to existing literature. Throughout this paper when we refer as $\tilde{O}(n^{\frac{1}{2k+3}})$ as optimal regret we assume that $C_n = n^k\|D^{k+1} \bs \theta_{1:n}\|_1$ is $O(1)$.

\begin{table*}[ht!]
\centering
\caption{\emph{Summary of regret bounds for \ADAVAW{} run with a fixed input parameter $k$ alongside bounds for various other policies. \ADAVAW{} is adaptively optimal for an array of distinct sequence classes featuring varying degrees of smoothness. We assume similar assumptions as in the description of Table \ref{tab:TV-rates}. We adopt the notation $a \wedge b = \min\{ a,b\}$.}}\label{tab:adaptive}
\resizebox{\columnwidth}{!}{
\begin{tabular}{|c|c|c|c|}
\hline
\textbf{Sequence Class}                                                                                                & \multicolumn{3}{c|}{\textbf{Dynamic Regret}} \\ \cline{2-4}
    &   \ADAVAW{} & ARROWS & MA/OGD/Ader\\ \hline
    \begin{tabular}{c} $TV^k(C_n):$\\$ n^k\|D^{k+1}\bs \theta_{1:n}\|_1 \le C_n$\end{tabular} & $\tilde O\left (n^{\frac{1}{2k+3}}C_n^{\frac{2}{2k+3}} \right)$ & $\tilde{O}\left(n^{1/3}C_n^{2/3}\right)$ & $\tilde O\left(\sqrt{nC_n}\right)$\\ \cline{1-4}
    \tstrut 
 \begin{tabular}{c}$\cS^{k+1}\left(\frac{C_n}{\sqrt{n}} \right):$\\ $n^{k}\|D^{k+1}\bs \theta_{1:n}\|_2 \le \frac{C_n}{\sqrt{n}}$\end{tabular}  \bstrut &        $\tilde O\left (n^{\frac{1}{2k+3}}C_n^{\frac{2}{2k+3}} \right)$ & $\tilde{O}\left(n^{1/3}C_n^{2/3}\right)$ & $\tilde{O}\left(n^{1/3}C_n^{2/3}\right)$  \\  \cline{1-1}
    \tstrut \begin{tabular}{c}$
\cH^{k+1} \left(\frac{C_n}{n} \right):$\\ $n^k\|D^{k+1}\bs \theta_{1:n}\|_\infty \le \frac{C_n}{n}$\end{tabular}   \bstrut & & & \\ \hline
    \tstrut \begin{tabular}{c}$\mathcal{E}^{k+1}(J_n):$\\$\|D^{k+1} \bs \theta_{1:n}\|_0 \le J_n, \: J_n \ge 1$\end{tabular} & $\tilde O(J_n)$ & $\tilde{O}\left(n^{1/3}J_n^{2/3}\right)$ & $\tilde O\left( \sqrt{nJ_n} \right)$ \\ \hline
\end{tabular}
}
\end{table*}

\textbf{Non parametric Regression} As noted in Section \ref{sec:problem-setup}, the problem setup we consider in this paper can be regarded as an online version of the batch non parametric regression framework. It has been established (see for eg, \citep{locadapt,donoho1998minimax,tibshirani2014adaptive} that minimax rate for estimating sequences with bounded $TV^k$ distance under squared error loss scales as $n^{\frac{1}{2k+3}}(n^k\|D^{k+1} \bs \theta_{1:n}\|_1)^{\frac{2}{2k+3}}$ modulo logarithmic factors of $n$. However, the problem of forecasting is more challenging than the offline setup because while making a prediction, we do not see the noisy realizations of ground truth for the future time points. In this work we connect together several ideas from online learning and batch regression setting to achieve a $\tilde{O}\left(n^{\frac{1}{2k+3}}(n^k\|D^{k+1} \bs \theta_{1:n}\|_1)^{\frac{2}{2k+3}}\right)$ minimax dynamic regret for the forecasting problem.

\textbf{Non-stationary Stochastic Optimization} As mentioned before in Section \ref{sec:problem-setup}, our forecasting framework can be considered as a special case of non-stationary stochastic optimization setting studied in \citep{besbes2015non,chen2018non}. A path variational constraint $V_n:=\sum_{t=1}^{n-1} \|f_{t+1} - f_t \|_\infty$ is defined in \citep{besbes2015non}. With squared error losses $f_t(x) = (x - \theta_t)^2$ and the boundedness constraint on ground truth in Assumption (A5), it can be shown that $V_n = O(\|D \bs \theta_{1:n}\|_1)$. Then their proposed algorithm namely, Restarting Online Gradient Descend (OGD) yields a dynamic regret of $O\left(n^{1/2}(\|D\theta_{1:n}\|_1)^{1/2}\right)$ for our problem. Due to Proposition \ref{prop:lower}, we see that the rate wrt $n$ is suboptimal for TV orders $k \ge 0$. Finally to achieve this rate, restarting OGD requires the knowledge of a tight bound on $\|D \bs \theta_{1:n}\|_1$ ahead of time which may not be practical on all occasions. Similar conclusions can be drawn if we consider the work of \citep{chen2018non}.

\textbf{Prediction of Bounded Variation sequences} Our problem setup is identical to that of \citep{arrows2019} except for the fact that they consider forecasting sequences whose zeroth order Total Variation is bounded. Our work can be considered as a generalization to any TV order $k$. As the value of $k$ increases, the sequence becomes more regular and one expects sharper rates for dynamic regret. However the algorithm of \citep{arrows2019} gives a suboptimal regret of $O(n^{1/3})$ for $k \ge 1$ even when both $\|D \bs \theta_{1:n}\|_1$ and $n^k\|D^{k+1} \bs \theta_{1:n}\|_1$ are $O(1)$.

We enumerate the comprehensive list of differences of this work when compared to \citep{arrows2019} for quick reference.
\begin{itemize}
    \item We work with a strictly general path varaiational that promotes piecewise polynomial structure in the comparator sequence. The path variational in \citep{arrows2019} promotes piecewise constant structures.
    \item By exploiting connections to regression splines, we formulate a more general restarting rule than \citep{arrows2019}.
    \item We demonstrate that zero padding (and many other padding approaches) prior to computing wavelet transform as done in \citep{arrows2019} will not preserve the higher order total variation, thus lead to \emph{far sub-optimal} results for the current problem. We then propose a novel packing scheme to alleviate this.
    \item We exploit the structure of CDJV wavelets and present a significantly more involved analysis to obtain \emph{sharper} dynamic regret guarantees. Haar wavelets that worked in \citep{arrows2019}, did not work here.
    \item We characterise the optimality of our algorithm for the case of exact sparsity as done in Section \ref{sec:exact_sparse} which was not studied in \citep{arrows2019}. Sharper dynamic regret guarantees for higher order discrete Sobolev and Holder classes are also obtained.
    \item We extend the framework to prediction in higher dimensions (Remark \ref{rem:highdim}). We identify a class of loss functions other than squared error losses in which the dynamic regret guarantees of \ADAVAW{} still holds (Remark \ref{rem:genloss}).
\end{itemize}

To gain some perspective, we present a way to analyse the dynamic regret of existing strategies for our problem. Recall that due to \eqref{eqn:dyn-regret} the comparator sequence can be considered to be the ground truth $\bs \theta_{1:n}$. In the univariate setting, most of the existing dynamic regret bounds depends on the variational measure $\|D \bs \theta_{1:n} \|_1$. If we assume that first $k+1$ values of the sequence $\theta_{1:n}$ are zero, then by applying the inequality $\|D^{j-1} \bs \theta_{1:n}\|_1 \le n \|D^{j} \bs \theta_{1:n}\|_1$, starting at $j=k+1$ and proceeding iteratively towards $j=1$, we get $\|D \bs \theta_{1:n}\|_1 \le n^k \|D^{k+1} \bs \theta_{1:n}\|_1$. This will enable us to get regret bounds for algorithms whose dynamic regret depends on the quantity $\|D \bs \theta_{1:n}\|_1$. The bounds obtained in this manner is shown in Table \ref{tab:TV-rates}. 

Using similar arguments, it can be shown that $\cS^{k+1}(C_n/\sqrt{n}) \subseteq \cS^{1}(C_n/\sqrt{n})$ for bounded sequences. This results in the regret bounds for policies other than \ADAVAW{} as displayed in Table \ref{tab:adaptive} for Sobolev and Holder classes.

\textbf{Adaptive Online Learning} Our problem can also be cast as a special case of various dynamic regret minimization frameworks such as \citep{zinkevich2003online,hall2013dynamical,besbes2015non,chen2018non,jadbabaie2015online,hazan2007adaptive,daniely2015strongly,yang2016tracking,zhang2018adaptive,zhang2018dynamic,chen2018smoothed}. To the best of our knowledge, none of the algorithms presented in these works can achieve the optimal dynamic regret of ${O}(n^{\frac{1}{2k+3}})$. 

\textbf{Competitive Online Non parametric Regression} \citep{rakhlin2014online} considers an online learning framework with squared error losses where a sequence $y_1,\ldots,y_n$ is revealed by an adversary and the agent makes prediction $s_t$ at time $t$ that depends only on the past history. They only require the the sequence $\bs{y_{1:n}}$ to be coordinatewise bounded and no stochastic relations between ground truth and revealed labels are assumed. They consider a regret defined as,
\begin{align}
    R
    &:=  E\left[\sum_{t=1}^{n} (y_t - s_t)^2 - \inf_{f \in \cF} \sum_{t=1}^{n}(y_t - f(x_t))^2 \right], \label{eqn:comp-regret}
\end{align}
for a non parametric function class $\cF$. If we consider $\cF$ as the function class with bounded $TV^k$ distance, then their regret bounds implies an upperbound on the dynamic regret in \eqref{eqn:dyn-regret}. This can be seen by setting $f(x_t) = \bs \theta_{1:n}[t]$ for $\bs \theta_{1:n} \in TV^k(C_n)$ and $y_t = \bs \theta_{1:n}[t] + \epsilon_t$ for independent subgaussian $\epsilon_t$, $t=1,\ldots,n$. Then,
\begin{align}
    R
    &\ge  E\left[\sum_{t=1}^{n} (y_t - s_t)^2 -  \sum_{t=1}^{n}(y_t - \bs \theta_{1:n}[t])^2\right],\\
    &=_{(a)} \sum_{t=1}^{n} E[s_t^2] + 2E[y_t\bs \theta_{1:n}[t]] - (\bs \theta_{1:n}[t])^2 - 2E[y_t]E[s_t],\\
    &= E\left[\sum_{t=1}^n \left(s_t- \bs \theta_{1:n}[t]\right)^2 \right],
\end{align}
where (a) is follows from the fact that the forecaster's prediction $s_t$ is independent of $y_t$.

The results of \citep{rakhlin2014online} on Besov spaces with squared error loss establishes that minimax rate for the online setting for the problem at hand is also same as that of the iid batch setting. They prove that minimax rate for Besov spaces indexed by $B_{p,q}^s$ is $O(n^{1/(2s+1)})$ in the univariate case whenever $s \ge 1/2$. The $TV^k(C_n)$ class is sandwiched between two Besov spaces $B_{1,1}^{k+1}$ and $B_{1,\infty}^{k+1}$ for an appropriate scaling of the radius. Since the two Besov spaces has the same minimax rate, the minimax dynamic regret for forecasting $TV^{k}(C_n)$ sequences in the online setting is also $O(n^{1/2k+3})$. However, the arguments in \citep{online_nonpar2015} are non-constructive. They propose a generic recipe based on relaxations of sequential Rademacher complexity for designing optimal online policies. However, we were unable to come up with a relaxation that can lead to computationally tractable forecasters that has the optimal dependence of $n$ and variational budget $\|D^{k+1} \bs \theta_{1:n}\|_1$ on the regret rate.

\citep{gaillard2015chaining} proposes a chaining algorithm to optimally control \eqref{eqn:comp-regret} when $\cF$ is taken to be the class of Holder smooth functions. Consequently, their algorithm yields optimal rates for dynamic regret defined in \eqref{eqn:dyn-regret} when $\theta_t$ are samples of a Holder smooth function. Such functions are spatially more regular than those present in a TV ball. In section 6.2, we show that our proposed policy \ADAVAW{} achieves the  optimal dynamic regret for Holder spaces enclosed within a higher order TV ball with faster run time complexity. 

Other works that can be cast under the setting described in \citep{rakhlin2014online} such as \citep{kotlowski2016, koolen2015minimax} are all unable to achieve the optimal dynamic regret for the problem at hand.

\textbf{Classical Time Series Forecasters} Algorithms such as ARMA \citep{box1970time} and Hidden Markov Models \citep{baum1966statistical} aims to detect recurrent patterns in a stationary stochastic process. However, we focus on surfacing out the hidden trends in a non-stationary stochastic process. Our work is more closely related to the idea of Trend Smoothing, similar in spirit to that of Hodrick-Prescott filter \citep{hodrick1997postwar} and \citep{l1tf}.

\textbf{Exact Sparsity} It is established in \citep{guntuboyina2018constrainedTF} that Trend Filtering can achieve a total squared error rate of $\tilde{O}(J_n)$ for $\cE^{k+1}(J_n)$ (defined in Section \ref{sec:exact_sparse}) in the batch setting. In each of the $J_n$ stable sections, the gradient of the polynomial signal is zero atmost $k$ times. With the boundedness assumption this yields a TV0 distance atmost $B(k+1)$ within a single section. At the change points the TV0 distance encountered is atmost $B$. Summing across all $J_n$ sections yields a total TV0 distance of $O(KJ_n)$. This bound on TV0 distance can be used to derive the rates of $O(n^{1/3}J_n^{2/3})$ for ARROWS \citep{arrows2019} and $O(\sqrt{nJ_n})$ for policies presented in \citep{besbes2015non,chen2018non,zinkevich2003online,zhang2018adaptive}. (See Table \ref{tab:adaptive})


\section{Analysis}
\subsection{Connecting wavelet coefficients and higher order $TV^k$ distance}
\begin{lemma} \label{lem:tvlb}
Let $\tilde {\bs \theta}_{1:t} = \texttt{recenter}({\bs \theta}_{1:t})$ and $(\bs a, \bs b) = \texttt{pack}(\tilde {\bs \theta}_{1:t})$. For an orthonormal DWT matrix $\bs W$,
\begin{align}
    \frac{\|\bs{W a} \|_2 + \| \bs{Wb}\|_2}{\sqrt{t}}
    &\lesssim t^{k}\|D^{k+1} {\bs \theta}_{1:t}\|_1,
\end{align}
where we have subsumed constants that depend only on $k$.
\end{lemma}
\begin{proof}
Consider the truncated power basis with knots at points $\frac{1}{n},\frac{2}{n},\ldots,1$ defined as follows:
\begin{align}
    g_1(x) = 1, \: g_2(x) = x, \ldots,\: g_k(x) = x^k\\
    g_{k+1+j}(x) = \left(x-\frac{j}{n}\right)_+^k, \: j=1,\ldots,n-k-1,
\end{align}
$x_+ = \max\{x,0\}$. Since an $t \times t$ matrix $\bs G$ with entries $g_{j}(\frac{i}{t})$ at the position $(i,j)$ is invertible, we can write any sequence ${\bs \theta}_{1:t}$ as
\begin{align}
    {\bs \theta}_{1:t}[i] = \sum_{j=1}^{t} \beta_j g_j(\frac{i}{t}),
\end{align}
for $i=1,\ldots,t$. From the above equation we see that,

\begin{align}
    t^{k}\|D^{k+1} {\bs \theta}_{1:t} \|_1
    &= k! \sum_{j=k+2}^{t} |\beta_j| \label{eqn:Dtheta}
\end{align}

Let $\tilde {\bs \theta}_{1:t} = \texttt{recenter}({\bs \theta}_{1:t})$. Let $\tilde{\bs g}_{j} = \texttt{recenter}(\bs g_{j})$ where $\tilde{\bs g}_j$ is the $j^{th}$ column of the matrix $\bs G$. Since $\|\bs g_j\|_\infty \le 1$ we have $\|\tilde{\bs g}_{j}\|_\infty = O(1)$ where the hidden constant only depends on $k$.

Thus
\begin{align}
    \|\tilde {\bs \theta}_{1:t}\|_\infty
    &= \left \| \sum_{j=k+2}^{t} \beta_j \tilde{\bs g}_{j} \right\|_\infty,\\
    &\le \sup_{k+2 \le i \le t} \|\tilde{\bs g}_{i}\|_\infty \sum_{j=k+2}^{t} |\beta_j|,\\
    &\lesssim t^{k}\|D^{k+1} {\bs \theta}_{1:t} \|_1, \label{eqn:pad-bound}
\end{align}
where the last line follows from \eqref{eqn:Dtheta}. We subsume a constant that only depends on $k$. Now using $\|\bs x\|_2 \le \sqrt{m} \| \bs x\|_\infty$  for $\bs x \in \mathbb{R}^m$, we have
\begin{align}
    \frac{\|\tilde {\bs \theta}_{1:t}\|_2}{\sqrt{t}}
    &\lesssim t^{k}\|D^{k+1} {\bs \theta}_{1:t} \|_1.
\end{align}

We have thus established a lower-bound on the TV using the energy of the OLS residuals. For a vector $\bs z$ let $(\bs x, \bs y) = \texttt{pack}(\bs z)$. We have the following relations,

\begin{align}
    \|\bs z\|_2
    &\ge \sqrt{\frac{\|\bs x\|_2^2 + \|\bs y\|_2^2}{2}},\\
    &\ge \frac{\|\bs x\|_2 + \|\bs y\|_2}{2},
\end{align}
where the last line follows from Jensen's inequality and the concavity of $\sqrt{\cdot}$ function.

\end{proof}

\subsection{Bounding the Regret} \label{app:regret}

Our proof strategy falls through the following steps.

\begin{enumerate}
    \item Obtain a high probability bound of bias variance decomposition type on the total squared error incurred by the policy within a bin.
    \item Bound the variance by optimally bounding the number of bins spawned.
    \item Bound the bias using the restart criterion and adaptive minimaxity of soft-thresholding estimator \citep{donoho1998minimax}.
\end{enumerate}

\begin{lemma} \label{lem:bias-variance bound}
(\textbf{bias-variance bound)})
Let $E[\hat y_t] = p_t$. For any bin $[t_h,t_l]$ with $t_h \ge k$ discovered by the policy, we have with probability atleast $1-\delta/2$
\begin{align}
\sum_{t=t_h}^{t_l} (\hat{y}_t - {\bs \theta}_{1:n}[t])^2
&\le \sum_{t=t_h}^{\bar t_l}  2(p_t - {\bs \theta}_{1:n}[t])^2 + 4 \sigma^2 (k+1) \log\left(1+\frac{n^{2k+3}}{k+1}\right)\log (4n^3/\delta).
\end{align}
\end{lemma}
\begin{proof}
First let's consider an arbitrary interval $[\ubar l, \bar l]$ such that $\ubar l \ge k$. We proceed to bound the bias and variance of predictions made by a VAW forecaster. Note that the bin $[\ubar l, \bar l]$ is arbitrary and may not be an interval discovered by the policy. The predictions made by VAW forecaster at time $t \in [\ubar l, \bar l]$ is given by,
\begin{align}
    \hat y_t
    &= \langle x_t, \tilde{\bs{A_t}}^{-1} \sum_{s=\ubar l-k}^{t-1} y_s \bs{x_s}\rangle,
\end{align}
where $\tilde{\bs{A_t}} = \bs{I}+\sum_{s=\ubar l-k}^{t} \bs{x_s}\bs{x_s}^T$.

Let
\begin{align}
    p_t
    &= E[\hat{y}_t],\\
    &= \langle x_t, \tilde{\bs{A_t}}^{-1} \sum_{s=\ubar l-k}^{t-1} {\bs \theta}_{1:n}[s \bs{x_s}\rangle.
\end{align}

For notational convenience, define 
\begin{align}
    \bs{X_t} = [\bs{x_{\ubar l -k}},\ldots,\bs{x_t}]^T. \label{eq:designmatrix}
\end{align}

Let
\begin{align}
    \Var{(\hat{y}_t)}
    &= \sigma^2 \bs{x_t}^T \tilde{\bs{A_t}}^{-1}\bs{X_t}^T\bs{X_t}\tilde{\bs{A_t}}^{-1}\bs{x_t},\\
    &\le \sigma^2 \bs{x_t}^T \tilde{\bs{A_t}}^{-1}\bs{x_t},\\
    &= \sigma_t^2
\end{align}
where the last line is due to $\bs{X_t}^T\bs{X_t} \preccurlyeq \tilde{\bs{A_t}}$, where $\bs{U} \preccurlyeq  \bs{V}$ means $\bs V- \bs U$ is a Positive Semi Definite matrix.

Define a normalized random variable
\begin{align}
    \label{eq:normrv}
    Z_t = \frac{\hat{y}_t - p_t}{\sigma_t}.
\end{align}

Thus $Z_t$ is a sub-gaussian random variable with variance parameter 1. By sub-gaussian tail inequality we have,
\begin{align}
    P\left( |Z_t| \ge \sqrt{2\log (4n^3/\delta)} \right) \le \delta/2n^3,
\end{align}
for some $\delta \in (0,1]$. Noting that length of a bin is atmost $n$, an application of uniform bound yields

\begin{align}
    P\left(\sup_{\ubar l \le t \le l} |Z_t| \ge \sqrt{2\log (4n^3/\delta)} \right) \le \delta/2n^2.
\end{align}

Adding and subtracting a ${\bs \theta}_{1:n}[t]$ to the numerator of \eqref{eq:normrv}, we get that with probability atleast $1-\delta/2n^2$,
\begin{align}
    | \hat{y}_t - {\bs \theta}_{1:n}[t] |
    &\le | p_t - {\bs \theta}_{1:n}[t] | + \sigma_t\sqrt{2\log (4n^3/\delta)}, \forall t \in [\ubar l, \bar l].
\end{align}

Hence the squared error within a bin can be bounded in probability as
\begin{align}
    \sum_{t=\ubar l}^{\bar l} (\hat{y}_t - {\bs \theta}_{1:n}[t])^2
    &\le  \sum_{t=\ubar l}^{\bar l}  2(p_t - {\bs \theta}_{1:n}[t])^2 + 4\sigma_t^2 \log (4n^3/\delta), \label{eq:hp1}
\end{align}
where we used $(a+b)^2 \le 2a^2 + 2b^2$. 

Let's focus on the second term in \eqref{eq:hp1}. By lemma 11.11 of \citep{BianchiBook2006} and by following the arguments of proof of Theorem 11.7 there, we get
\begin{align}
    \sum_{t=\ubar l}^{\bar l} \sigma_t^2
    &\le \sigma^2 \sum_{d=1}^{k+1} \log(1+\lambda_d),
\end{align}
where $\lambda_d$ are the eigenvalues of the $(k+1)\times (k+1)$ matrix $\tilde{\bs{A}}_{\bar l} - \bs{I}$. It is well known that $\tilde{\bs{A}}_{\bar l} - \bs{I}$ has the same nonzero eigenvalues as the Gram matrix $\bs{G}$ with entries $G_{i.j} = \bs{x_i}^T\bs{x_j}$. Note that $\|\bs{x_t}\|_2^2 \le n^{2k+2}, \forall t \in [1,n]$. Since the product $\Pi_{d=1}^{k+1} (1+\lambda_d)$ is maximised when $\lambda_d = (\ubar l - \bar l) n^{2k+2}/(k+1) \le n^{2k+3}/(k+1)$ we have,
\begin{align}
    \sigma^2 \sum_{d=1}^{k+1} \log(1+\lambda_d)
    &\le \sigma^2 (k+1) \log(1+\frac{n^{2k+3}}{k+1}). \label{eqn:matdet}
\end{align}

Thus with probability atleast $1-\delta/n^2$
\begin{align}
    \sum_{t=\ubar l}^{\bar l} (\hat{y}_t - {\bs \theta}_{1:n}[t])^2 
    &\le  \sum_{t=\ubar l}^{\bar l}  2(p_t - {\bs \theta}_{1:n}[t])^2 + 4 \sigma^2 (k+1) \log \left(1+\frac{n^{2k+3}}{k+1}\right)\log (4n^3/\delta). 
\end{align}

As mentioned earlier, the bin $[\ubar l, \bar l]$ can be arbitrary and may not be discovered by policy. However, we want to analyze the Total Squared Error (TSE) incurred within true bins spawned by the policy. A small caveat here is that observations within such true bins satisfy the restart criteria and can't be regarded as independent random variables. To get rid of this problem, we use a uniform bound argument to bound the TSE incurred in all possible $O(n^2)$ bins. This leads to
\begin{align}
    P \left(\sup_{[\ubar l, \bar l]} \sum_{t=\ubar l}^{\bar l} (\hat{y}_t - {\bs \theta}_{1:n}[t])^2 - \sum_{t=\ubar l}^{\bar l}  2(p_t - {\bs \theta}_{1:n}[t])^2 - 4 \sigma^2 (k+1) \log \left(1+\frac{n^{2k+3}}{k+1}\right)\log (4n^3/\delta) \ge 0 \right) \le \delta/2.
\end{align}
\end{proof}

\begin{lemma} \label{lem:subgaussian_coeffs}
(\textbf{subgaussian wavelet coefficients})
Let $(\bs{y_1},\bs{y_2}) = \texttt{pack}\left(\texttt{recenter}(\textbf{y})\right)$ for a vector $\bs y$ of observations of length $L$. Let $(\bs{\alpha_1},\bs{\alpha_2}) = (\bs{Wy_1},\bs{Wy_2})$ for an orthonormal DWT matrix $\bs{W}$. Then both $\bs{\alpha_1}$ and $\bs{\alpha_2}$ are marginally subgaussian with parameter $4\sigma^2$.
\end{lemma}
\begin{proof}
From the theory of least squares regression,
\begin{align}
    \texttt{recenter}(\bs{y}) 
    &= \bs{y} - \bs{X_L}(\bs{X_L}^T \bs{X_L})^{-1}\bs{X_L}^T \bs{y},
\end{align}
where $\bs{X_L}$ is defined as in \eqref{eq:designmatrix}. Since $L \ge k+1$, $\bs{X_L}^T \bs{X_L}$ can be shown to be invertible. (see for eg. lemma \ref{lem:det_poly})

Without loss of generality, we proceed to characterize the sub-gaussian behaviour of the \emph{first} wavelet coefficient of $\bs{y_1}$. The extension to other wavelet coefficients is straight forward.

Let $\bs u^T$ be the first row of the wavelet transform matrix $\bs W$ whose dimension is compatible to $\bs{y_1}$. Let's augment $\bs u^T$ as follows.
\begin{align}
    \tilde{\bs u}^T
    &= [\bs u^T, \bs 0^T],
\end{align}
such that length of $\tilde{\bs u}$ is $L$.

We have,
\begin{align}
    \bs{\alpha_1}[0] 
    &= \tilde{\bs u}^Ty - \tilde{\bs u}^T \bs{X_L}(\bs{X_L}^T \bs{X_L})^{-1}\bs{X_L}^T \bs{y}. \label{eq:coeff}
\end{align}

\eqref{eq:coeff} along with noisy feedback  implies that $\bs{\alpha_1}[0]$ is a Lipschitz function of $L$ iid subgaussian random variables. Then by Proposition 2.12 from \citep{DJBook},  $\bs{\alpha_1}[0]$ is also subgaussian with variance parameter given by the square of Lipschitz constant $\ell^2$ times $\sigma^2$. Since $\bs{\alpha_1}[0]$ is a linear function of the iid subgaussians we have,

\begin{align}
    \ell
    &= \|\tilde{\bs u} -  \bs{X_L}(\bs{X_L}^T \bs{X_L})^{-1}\bs{X_L}^T \tilde{\bs u}\|_2,\\
    &\le \|\tilde{\bs u}\|_2 + \|\bs{X_L}(\bs{X_L}^T \bs{X_L})^{-1}\bs{X_L}^T \tilde{\bs u}\|_2,\\
    &\le_{(a)} \|\bs{u}\|_2 + \|\bs{X_L}(\bs{X_L}^T \bs{X_L})^{-1}\bs{X_L}^T\|_2 \|\bs{u}\|_2,\\
    &=_{(b)} 2.
\end{align}
In (a) we used $\|\bs{Ax}\|_2 \le \|\bs A\|_2 \|\bs x\|_2$ where $\|\bs A\|_2$ is the induced matrix norm and the fact that $\|\tilde{\bs u}\|_2 = \|\bs{u}\|_2$. In (b) we notice that $ \|\bs{u}\|_2 = 1$ as the DWT matrix $\bs W$ is orthonormal and $\|\bs{X_L}(\bs{X_L}^T \bs{X_L})^{-1}\bs{X_L}^T\|_2 = 1$ since $\bs{X_L}(\bs{X_L}^T \bs{X_L})^{-1}\bs{X_L}^T$ is a projection matrix.

Similarly it can be shown that $\bs{\alpha_2}$ is marginally subgaussian with parameter $4\sigma^2$.
\end{proof}

\begin{lemma} \label{lem:uni-shrinkage}
(\textbf{uniform shrinkage})
Assume the setting of lemma \ref{lem:subgaussian_coeffs}. Let $(\bs{\hat{\alpha_1}},\bs{\hat{\alpha_2}}) = (T(\bs{\alpha_1}),T(\bs{\alpha_2}))$ where $T(\cdot)$ is the soft-thresholding operator with threshold $\sigma\sqrt{\beta \log n}$. Then with probability atleast $1-2n^{3-\beta/8}$, $|(\bs{\hat{\alpha_r}})_i| \le |E\left[(\bs{\alpha_r})_i\right]|$ for each co-ordinate $i$ and $r = 1,2$. The expectation is taken wrt to randomness in the observations.
\end{lemma}
\begin{proof}
Consider a fixed bin $[\ubar l, \bar l]$. Due to results of lemma \ref{lem:subgaussian_coeffs} and subgaussian tail inequality,
\begin{align}
 P\left(|(\bs{\hat{\alpha_r}})_i - E\left[(\bs{\alpha_r})_i\right]| \ge \sigma\sqrt{\beta \log n}\right) 
 &\le 2n^{-\beta/8}.   
\end{align}

Then arguing in the similar lines as in the proof of lemma 15 of \cite{arrows2019}, the result follows.
\end{proof}

\begin{lemma} \label{lem:bins}
\textbf{(bin control) }
With probability atleast $1-2n^{3-\beta/8}$, the number of bins $M$, spawned by the policy is atmost\\
$\min\left\{n,\max\{1,\tilde{O}(n^{\frac{1}{2k+3}}\|n^kD^{(k+1)}\bs{\theta}_{1:n}\|_1^{\frac{2}{2k+3}})\}\right\}$ where $\tilde{O}$ hides factors that depend on wavelet function, constants that only depend on TV order $k$ and polynomial factors of $\log n$.
\end{lemma}
\begin{proof}
Let $L_i$ be the length of the $i^{th}$ bin. Let $\hat{\bs{\alpha}}_{1i}, \hat{\bs{\alpha}}_{2i}$ be the denoised wavelet coefficient segments of the re-centered observations within a bin $i$ as described in the policy and $\bs{\theta_i}$ be the ground truth vector in bin $i$.

By the policy's restart rule,
\begin{align}
    \frac{\sigma}{\sqrt{L_i}}
    &\le \frac{1}{\sqrt{L_i}}\left( \|\hat{\bs{\alpha}}_{1i}\|_2 +  \|\hat{\bs{\alpha}}_{2i}\|_2\right).
\end{align}

Due to the uniform shrinkage property specified in lemma \ref{lem:uni-shrinkage}, we have with probability atleast $1-2n^{3-\beta/8}$
\begin{align}
    \frac{\sigma}{\sqrt{L_i}}
    &\le \frac{1}{\sqrt{L_i}}\left( \|\bs{\alpha}_{1i}\|_2 +  \|\bs{\alpha}_{2i}\|_2\right),\\ 
    &\lesssim_{(a)} 2^k L_i^k \|D^{k+1}\bs{\theta_i}\|_1,
\end{align}
where (a) follows due to lemma \ref{lem:tvlb}. The factor of $2^k$ is due to the fact that length of vectors $\bs{\alpha}_{1i}$ or $\bs{\alpha}_{2i}$ is atmost $2L_i$. The last line implies that when the $TV^k$ distance is zero, \ADAVAW{} doesn't restart with high probability making $M = 1$.

Rearranging and summing across all bins yields
\begin{align}
    \sum_{i=1}^{M} \frac{\sigma}{L_i^{k+1/2}}
    &\lesssim \|D^{k+1}\bs{\theta}_{1:n}[t]\|_1.
\end{align}

Now applying Jensen's inequality for the convex function $f(x) = \frac{1}{x^{k+1/2}}, x>0$, we get
\begin{align}
    \sigma M^{\frac{2k+3}{2}} n^{\frac{-(2k+1)}{2}}
    &\lesssim \|D^{k+1}\bs{\theta}_{1:n}\|_1,
\end{align}
where $\lesssim$ subsumes constants that depend only on wavelet functions, TV order $k$ and polynomial factors of $\log n$.

Rearranging the last expression yields the lemma.
\end{proof}

\begin{lemma}
\label{lem:vaw}
\textbf{(Vovk-Azoury-Warmuth regret)}
If the Vovk-Azoury-Warmuth forecaster with output denoted by $\hat v_j$ at time $j$, is run on a sequence\\ $(\boldsymbol{w_1},v_1),\ldots,(\boldsymbol{w_n},v_n) \in \mathbb{R}^{k+1} \times \mathbb{R}$, then for all $\boldsymbol{u} \in \mathbb{R}^{k+1}$,
\begin{align}
    \sum_{j=1}^{t} (\hat v_j - v_j)^2 - (\boldsymbol{u}^T \bs{w_j} - v_j)^2
    &\le \frac{1}{2}\|\boldsymbol{u}\|_2^2 + \frac{(k+1)B^2}{2} \log\left(1+\frac{t^{k+2}}{k+1}\right),\\
    &= \tilde{O}(B^2),
\end{align}
where $B = \max_{i=1,\ldots,t}|y_i|$ and $\boldsymbol{w_j} = [1,j,\ldots,j^k]^T$.
\end{lemma}
\begin{proof}
The first inequality is due to Theorem 11.8 of \citep{BianchiBook2006}. The second equality follows because under the given choice of monomial features, it is shown in Corollary \ref{cor:bounded_OLScoeffs} that when $\bs{u}$ is the coefficient vector of OLS fit, $\|\bs{u}\|_2^2 = O(B^2)$.
\end{proof}

Next we characterize the optimality of soft-thresholding estimator on $TV^k$ class. The key to this is the Theorem 19 from \citep{arrows2019}.
\begin{theorem} \label{thm:soft-threshold_optimality}
\textbf{\citep{arrows2019}}
Consider the observation model $\breve{\bs{y}} = \breve{\bs \alpha} + \bs Z$, where $\breve{\bs{y}} \in \mathbb R^n$, $\bs Z$ is marginally subgaussian with parameter $\sigma^2$ and $\breve{\bs \alpha} \in \bs A$ for some solid and orthosymmetric $\bs A$. Let $\hat{\bs \alpha}_\delta$ be the soft thresholding estimator with input $\breve{\bs{y}}$ and threshold $\delta$. When $\delta = \sigma\sqrt{\beta \log n}$, with probability atleast $1-2n^{1-\beta/2}$ the estimator $\hat{\bs \alpha}_\delta$ satisfies
\begin{align}
      \|\hat{\alpha}_\delta - \alpha \|^2 
      &\le 8.88 \beta (1+\log(n)) \inf_{\hat{\alpha}}\sup_{\alpha\in A} E[\|\hat{\alpha}-\alpha\|^2].
\end{align}
\end{theorem}

We are interested in the case where $\bs A$ is the space of wavelet coefficients for $TV^k$ bounded fucntions. Since $TV^k$ class is sandwiched between two Besov spaces, it can be shown that $\bs A$ is solid and orthosymmetric (see for eg. \citep{DJBook}, section 4.8). Note that subtracting a polynomial of degree $k$ has no effect on the $TV^k$ distance. It has been established in lemma \ref{lem:subgaussian_coeffs} that OLS residual are subgaussian with parameter $4\sigma^2$. Hence we are under the observation model of Theorem \ref{thm:soft-threshold_optimality}. By the results of \citep{donoho1998minimax}, we have $\inf_{\hat{\alpha}}\sup_{\alpha\in A} E[\|\hat{\alpha}-\alpha\|^2] = \tilde{O}(n^{\frac{1}{2k+3}}(n^k D{\bs \theta}_{1:n}\|_1)^{\frac{2}{2k+3}}\sigma^{\frac{4k+4}{2k+3}})$. This along with using a uniform bound across all $O(n^2)$ bins leads to the following Corollary.

\begin{corollary} \label{cor:soft-threshold-optimality}
Under the observation model and notations in Theorem \ref{thm:soft-threshold_optimality} but with a subgassuan parameter $4\sigma^2$ when $\bs A$ is the wavelet coefficients of re-centered ground truth within a bin discovered by the policy, then with probability atleast $1-2n^{3-\beta/8}$
\begin{align}
     \|\hat{\alpha}_\delta - \alpha \|^2 
     &= \tilde{O}(n^{\frac{1}{2k+3}}(n^k D{\bs \theta}_{1:n}\|_1)^{\frac{2}{2k+3}}\sigma^{\frac{4k+4}{2k+3}}).
\end{align}
\end{corollary}

\begin{lemma} \label{lem:bias}
\textbf{(bias control)}
Let $E[\hat y_t] = p_t$. For any bin $[t_h,t_l]$, $L = t_l - t_h$, with $t_h \ge k$ discovered by the policy, we have with probability atleast $1-2n^{3-\beta/8}$
\begin{align}
\sum_{t=t_h}^{\bar t_l-1}  (p_t - {\bs \theta}_{1:n}[t])^2
&= \tilde{O}(1) + \tilde{O}\left(L^{\frac{2k+1}{2k+3}}\|D^{k+1}{\bs \theta}_{t_h-k:t_l-1}\|_1^{\frac{2}{2k+3}}\right) + (p_{t_l} - {\bs \theta}_{1:n}[t_l])^2.
\end{align}
\end{lemma}
\begin{proof}
For a bin $[t_h,t_l]$ let
\begin{align}
    T
    &=\sum_{t=t_h}^{t_l}  (p_t - {\bs \theta}_{1:n}[t])^2.
\end{align}

Note that $T$ is the squared error incurred by the VAW forecaster when run with the sequence ${\bs \theta}_{t_h:t_l}$. Let $\bs u$ be the coefficient of the OLS fit using monomial features for the ground truth $[{\bs \theta}_{t_h-k:t_l-1}]$. Further let's recall/adopt the following notations:
\begin{enumerate}
    \item[1] $(\bs{g_1},\bs{g_2}) = \texttt{pack}\left(\texttt{recenter}({\bs \theta}_{t_h-k:t_l-1})\right)$;
    \item[2] $(\bs{\alpha_1},\bs{\alpha_2}) = (\bs{Wg_1},\bs{Wg_2})$;
    \item\citep{arrows2019} $(\bs{y_1},\bs{y_2}) = \texttt{pack}\left(\texttt{recenter}(\bs{y}_{t_h-k:t_l-1})\right)$;
    \item[4] $L = t_l-t_h+k$;
    \item[5] $(\hat{\bs{\alpha_1}},\hat{\bs{\alpha_2}}) = (T(\bs{Wy_1}),T(\bs{Wy_2}))$ where $T(\cdot)$ is soft-thresholding operator at threshold $\sigma \sqrt{\beta \log n}$.
\end{enumerate}

\begin{align}
    T - (p_{t_l} - {\bs \theta}_{1:n}[t_l])^2
    &\le_{(a)} \sum_{j=t_h-k}^{t_l-1} (\bs{u}^T\bs{x_j} - {\bs \theta}_{1:n}[j])^2 + \tilde{O}(B^2),\\
    &\le_{(b)} \|\bs{\alpha_1}\|_2^2 + \|\bs{\alpha_2}\|_2^2 + \tilde{O}(B^2),\\
    &\le_{(c)} \|\hat{\bs{\alpha_1}}\|_2^2 + \|\hat{\bs{\alpha_2}}\|_2^2 + \|\hat{\bs{\alpha_1}} - \bs{\alpha_1}\|_2^2 + \|\hat{\bs{\alpha_2}} - \bs{\alpha_2}\|_2^2 + \tilde{O}(B^2) ,\\
    &\le_{(d)} \|\hat{\bs{\alpha_1}}\|_2^2 + \|\hat{\bs{\alpha_2}}\|_2^2 + \tilde{O}\left(L^{\frac{2k+1}{2k+3}}\|D^{k+1}{\bs \theta}_{t_h-k:t_l-1}\|_1^{\frac{2}{2k+3}}\sigma^{\frac{4k+4}{2k+3}}\right) + \tilde{O}(B^2),\\
    &\le_{(e)} \frac{\sigma^2}{L} + \tilde{O}\left(L^{\frac{2k+1}{2k+3}}\|D^{k+1}{\bs \theta}_{t_h-k:t_l-1}\|_1^{\frac{2}{2k+3}}\sigma^{\frac{4k+4}{2k+3}}\right) + \tilde{O}(B^2),\\
    &= \tilde{O}(1) +  \tilde{O}\left(L^{\frac{2k+1}{2k+3}}\|D^{k+1}{\bs \theta}_{t_h-k:t_l-1}\|_1^{\frac{2}{2k+3}}\right),
\end{align}

with probability atleast $1-2n^{3-\beta/8}$. Inequality (a) is due to lemma \ref{lem:vaw}, (b) is due to orthonormality of wavelet transform matrix $\bs W$, (c) by triangle inequality, (d) by Corollary \ref{cor:soft-threshold-optimality} and (e) is due to the fact that restart condition is not satisfied in the interior of a bin.
\end{proof}

\thmMain*
\begin{proof}
Let $L_i$ be the length of the $i^{th}$ bin $[t_h^{(i)},t_l^{(i)}]$ discovered by the policy. Let
\begin{align}
    T_i
    &=\sum_{t=t_h^{(i)}}^{t_l^{(i)}}  (p_t - {\bs \theta}_{1:n}[t])^2.
\end{align}

From lemma \ref{lem:bias} we have with with probability atleast $1-2n^{3-\beta/8}$,
\begin{align}
    T_i
    &= \tilde{O}(1) + \tilde{O}\left(L_i^{\frac{2k+1}{2k+3}}\|D^{k+1}{\bs \theta}_{t_h^{(i)}-k:t_l^{(i)}-1}\|_1^{\frac{2}{2k+3}}\right) + (p_{t_l^{(i)}} - {\bs \theta}_{1:n}[t_l^{(i)}])^2\\
    &= \tilde{O}(1) + \tilde{O}\left(L_i^{\frac{2k+1}{2k+3}}\|D^{k+1}{\bs \theta}_{t_h^{(i)}-k:t_l^{(i)}-1}\|_1^{\frac{2}{2k+3}}\right),
\end{align}
where in the last line we used the fact that ground truths are bounded by $B$.

Now summing the squared bias across all $M$ bins discovered by the policy yields
\begin{align}
    T
    &= \sum_{i=1}^{M} T_i,\\
    &=_{(a)} \tilde{O(M)} + \sum_{i=1}^{M}\tilde{O}\left(L_i^{\frac{2k+1}{2k+3}}\|D^{k+1}{\bs \theta}_{t_h^{(i)}-k:t_l^{(i)}-1}\|_1^{\frac{2}{2k+3}}\right),\\
    &=_{(b)} \tilde{O}\left(n^{\frac{1}{2k+3}}\|n^kD^{(k+1)}\bs{\theta_{1:n}}\|_1^{\frac{2}{2k+3}}\right) + \sum_{i=1}^{M}\tilde{O}\left(L_i^{\frac{2k+1}{2k+3}}\|D^{k+1}{\bs \theta}_{t_h^{(i)}-k:t_l^{(i)}-1}\|_1^{\frac{2}{2k+3}}\right),\\
    &=_{(c)} \tilde{O}\left(n^{\frac{1}{2k+3}}\|n^kD^{(k+1)}\bs{\theta_{1:n}}\|_1^{\frac{2}{2k+3}}\right) +  \tilde{O}\left( \left(\sum_{i=1}^{M} L_i \right)^{\frac{2k+1}{2k+3}} \cdot \left(\sum_{i=1}^{M} \|D^{k+1}{\bs \theta}_{t_h^{(i)}-k:t_l^{(i)}-1}\|_1 \right)^{\frac{2}{2k+3}} \right),\\
    &= \tilde{O}\left(n^{\frac{1}{2k+3}}\|n^kD^{(k+1)}\bs{\theta_{1:n}}\|_1^{\frac{2}{2k+3}}\right) + \tilde{O}\left(n^{\frac{1}{2k+3}}\|n^kD^{(k+1)}\bs{\theta_{1:n}}\|_1^{\frac{2}{2k+3}}\right) \label{eqn:ub1},    
\end{align}

with probability atleast $1-4n^{3-\beta/8}$. Line (a) holds with probability atleast $1-2n^{3-\beta/8}$.  For (b) we used lemma \ref{lem:bins} and it holds with probability atleast $\left(1-2n^{3-\beta/8}\right)^2 \ge 1 - 4n^{3-\beta/8}$ . For (c) we used Holder's inequality $\bs x^T \bs y \le \|\bs x\|_p \|\bs y \|_q$ with $p = \frac{2k+3}{2k+1}$ and $q=\frac{2k+3}{2}$.

Since the variance within a bin is $\tilde{O}(\sigma^2)$ as indicated by lemma \ref{lem:bias-variance bound}, when summed across all bins we get a total variance of $\tilde{O}(\sigma^2 M)$ which is $\tilde{O}\left(n^{\frac{1}{2k+3}}\|n^kD^{(k+1)}\bs{\theta_{1:n}}\|_1^{\frac{2}{2k+3}}\right)$ by lemma \ref{lem:bins}.

A trivial upperbound for $T$ is
\begin{align}
    T
    &\le n(B^2 + \sigma^2),\\
    &= O(n). \label{eqn:ub2}
\end{align}

Combining \eqref{eqn:ub1} \eqref{eqn:ub2} and the variance summed across all terms yields
\begin{align}
    T
    &= \tilde{O}\left(\max \left\{n,n^{\frac{1}{2k+3}}\|n^kD^{(k+1)}\bs{\theta_{1:n}}\|_1^{\frac{2}{2k+3}}\right\}\right),
\end{align}
with probability atleast $1-4n^{3-\beta/8}-\delta/2$ where the dependence of $\delta$ in the failure probability is due to that fact that bias variance decomposition in lemma \ref{lem:bias-variance bound} holds with probability atleast $1-\delta/2$. By setting $\beta = 24+\frac{8\log(8/\delta)}{\log(n)}$, we get the Theorem \ref{thm:main}.
\end{proof}

\begin{remark} (\emph{Specialization to $k=0$})
When specialized to the case $k=0$, we recover the optimal rate established in \citep{arrows2019} for the bounded ground truth setting upto constants $B$ and $\sigma$. When $k=0$, our policy predicts $\frac{y_{t_h}+\ldots+y_{t-1}}{t-t_h+2}$ at time $t$. This is similar to online averaging except that the denominator is now $t-t_h+2$ instead of $t-t_h$. \citep{arrows2019} also considers the scenario where the point-wise bound on ground truth can increase in time as $O(C_n)$. As hinted by the similarity of \ADAVAW{} with that of \citep{arrows2019} for $k=0$ along with the fact that our restart rule also lower-bounds the Total Variation of ground truth with high probability, it is possible to get a regret bound of $\tilde{O}(n^{1/3}C_n^{2/3} + C_n^2)$ for \ADAVAW{} in this stronger setting. \end{remark}

\propLower*
\begin{proof}
Since a batch non-parametric regression algorithm is allowed to see the entire observations ahead of time, lower bound in the batch setting directly translates to lower bound for $R_{dynamic}$. Let $\cA_B$ be the set of all offline regression algorithms. The minimax rates of estimation of $TV^k$ bounded sequences under squared error losses from \citep{donoho1998minimax} gives,
\begin{align}
    \inf_{\bs s \in \cA_B} \sup_{{\bs \theta}_{1:n} \in TV^{(k)}(C_n)} \sum_{t=1}^{M} E\left[(\bs{s_t} - {\bs \theta}_{1:n}[t])^2 \right]\\
    = \Omega\left(n^{\frac{1}{2k+3}}C_n^{\frac{2}{2k+3}}\right).
\end{align}
\end{proof}

From \citep{donoho1990minimax}, minimax rates of estimation under squared error losses of sequences that satisfy $|{\bs \theta}_i| \le B$ scales as $\min\{nB^2,n\sigma^2\}$. Combining the two bounds yields Proposition \ref{prop:lower}. 

\propRuntime*
\begin{proof}
Let's describe the computational requirement at each time step. As outlined in Section 11.8 of \citep{BianchiBook2006}, we can use Sherman-Morrison formula to compute $A_t^{-1}$ in $O((k+1)^2)$ time. Using the same logic we can compute $(\bs{X_t}^T\bs{X_t})^{-1}$ needed by $\texttt{recenter}$ operation incrementally in $O((k+1)^2)$ time. Re-centering operation and computation of wavelet coefficients requires $O(n)$ time per round. Since there are $n$ rounds, the total run-time  complexity becomes $O((k+1)^2n^2)$.
\end{proof}

\textbf{Extension to higher dimensions} Consider a variational measure and the setup described in Remark \ref{rem:highdim}. Let $\hat{y}_t^{(i)}$ be the prediction of instance $i$ of \ADAVAW{} at time $t$. For each $i \in [d]$, we've
\begin{align}
    \sum_{t=1}^{n} (\hat{y}_t^{(i)} - \bs \theta_{1:n}[t][i])^2
    &= \tilde O \left(n^{\frac{1}{2k+3}}\Delta_i^{\frac{2}{2k+3}} \right),
\end{align}
by Theorem \ref{thm:main}. Summing across all dimensions yields,
\begin{align}
    R_n
    &= \sum_{i=1}^d \tilde O \left(n^{\frac{1}{2k+3}}\Delta_i^{\frac{2}{2k+3}} \right)\\
    &= \tilde{O} \left(d^{\frac{2k+1}{2k+3}} n^{\frac{1}{2k+3}} C_n^{\frac{2}{2k+3}} \right),
\end{align}
where the last inequality follows from applying Holder's inequality $\bs x^T \bs y \le \|\bs x\|_p \| \bs y \|_q$ to  $\sum_{i=1}^{d} 1^{\frac{2k+1}{2k+3}} \Delta_i^{\frac{2}{2k+3}}$ with norms $p = \frac{2k+3}{2k+1}$ and $q = \frac{2k+3}{2}$.

\textbf{Extension to general losses} Assume the interaction model in Figure \ref{fig:proto}. Instead of squared error losses, let the losses be $f_t$ as discussed in Remark \ref{rem:genloss}. Since $f_t$ is gamma smooth, we have
\begin{align} \label{eq:smooth}
    f_t(b) \le f_t(a) + f_t'(a) (b-a) + \frac{\gamma}{2} (b-a)^2.
\end{align}

Let $\hat{y}_t$ be the prediction of \ADAVAW{} at time t and $\bs \theta_t := \theta_{1:n}[t]$. Then regret with this loss function is
\begin{align}
    \sum_{t=1}^{n} f_t(\hat{y}_t) - f_t(\theta_t)
    &\le \sum_{t=1}^{n} \frac{\gamma}{2} (\hat{y}_t - \theta_t)^2,
\end{align}
by \eqref{eq:smooth} and using the fact $f_t'(\theta_t) = 0$. Now the statement in Remark \ref{rem:genloss} is immediate by appealing to Theorem \ref{thm:main}.

\subsection{Exact sparsity} \label{app:exact_sparse}
We start by the observation that an exact sparsity (i.e sparsity in the $\|\cdot\|_0$ sense) in the number of jumps of $\|D^{k+1}{\bs \theta}_{1:n}\|_0$ translates to an exact sparsity in the wavelet coefficients. This is made precise by the following lemma.

\begin{lemma} \label{lem:sparse-coefs}
Consider a sequence with $\|D^{k+1}{\bs \theta}_{1:n}\|_0 = J$. Then both the signals ${\bs \theta}_{1:n}$ and $\tilde {\bs \theta}_{1:n} = \texttt{recenter}({\bs \theta}_{1:n})$ can be represented using $O(k+J\log n)$ wavelet coefficients of a CDJV system of regularity $k+1$.
\end{lemma}
\begin{proof}
Throughout this proof when we say jumps, we refer to jumps in $\|D^{k+1}{\bs \theta}_{1:n}\|_0 $. Let $L = 2^{\ceil{\log_2 (k+1)}}$. Consider splitting the coefficients $\bs \alpha$ of the DWT transform into two parts: $\bs \alpha_{1:L}$ and $\bs \alpha_{L+1:n}$. By CDJV construction, the wavelets corresponding to indices $L+1,\ldots,n$ are all orthogonal to polynomials to degree atmost $k$. The space of polynomials of degree atmost $k$ is contained in the span of wavelets identified by the indices $1,\ldots,L$. Though the span of the first $L$ wavelets can also generate other waveforms which are not polynomials as well.

Notice that between two jumps, the underlying signal is a polynomial of degree atmost $k$. By orthogonality property discussed above, wavelet coefficients from the group $\bs \alpha_{L+1:n}$ assume the value zero if the support of corresponding wavelet is a region where the signal behaves as a polynomial. Since there are $J$ jump points and each point is covered by $\log n$ wavelets by the Multi Resolution property, there can be atmost $O(J\log n)$ non zero coefficients from the group $\bs \alpha_{L+1:n}$.

When we subtract the best polynomial fit due to the re-centering operation, it is only going to affect the first $L$ coefficients and keep the remaining unchanged. Hence the re-centered signal can have atmost $O(k + J\log n)$ nonzero coefficients.

\end{proof}

Due to lemmas \ref{lem:tvlb} and \ref{lem:uni-shrinkage}, the expression in the LHS of restart rule of the policy lower-bounds the $TV^k$ distance within a bin with high probability. So if a bin lies entirely between two jumps, we do not restart with high probability as the $TV^k$ distance is zero. This lead to the following Corollary.

\begin{corollary} \label{cor:sparse-restart}
Let $y_t = {\bs \theta}_t + \epsilon_t$, for $t=1,\ldots,n$ where $\epsilon_t$ are sub-gaussian with parameter $\sigma^2$ and $\|D^{k+1}{\bs \theta}_{1:n}\|_0 = J$ with $|{\bs \theta}_t| \le B$. Then with probability at-least $1-2n^{3-\beta/8}$ \ADAVAW{} restarts $O(J)$ times.
\end{corollary}

In the next Theorem, we characterize the optimality of soft-thresholding estimator in the exact sparsity case.

\begin{theorem} \label{thm:sparse-soft}
Under the setup of Corollary \ref{cor:sparse-restart}, the soft thresholding estimator whose estimates denoted by $\hat{\bs \alpha}_{1:n}$ with threshold set to $\sigma\sqrt{\log n}$ satisfy,
\begin{align}
    \|\hat{\bs \alpha}_{1:n} - {\bs \theta}_{1:n}\|_2^2
    &= \tilde{O}(J\sigma^2),
\end{align}
with probability atleast $1-2n^{1-\beta/2}$ where $\tilde{O}$ hides logarithmic factors of $n$.
\end{theorem}
\begin{proof}
Let $\bs \alpha$ denote the DWT coefficients of ${\bs \theta}_{1:n}$. By Gaussian tail inequality and union bound we have $P(\sup_{t}|\epsilon_t| \ge \sigma\sqrt{\log n}) \le 2n^{1-\beta/2}$. Conditioning on the event $\sup_{t}|\epsilon_t| \le \sigma\sqrt{\log n}$ we are under the observation model in lemma 17 of \cite{arrows2019}. Following the results there, with probability atleast $1-2n^{1-\beta/2}$ we have,
\begin{align}
    \|\hat{\bs \alpha}_{1:n} - {\bs \theta}_{1:n}\|_2^2
    &=\sum_{i=1}^{n} \min \left \{\bs \alpha[i]^2, 16\sigma^2 \log n\right \},\\
    &= \tilde{O}(J\sigma^2),
\end{align}
where the last line follows from lemma \ref{lem:sparse-coefs} and the fact that $O(k+J \log n) = O(KJ \log n) = O(J \log n)$.
\end{proof}

Now using a uniform bound argument across all $O(n^2)$ bins yields the following Corollary.
\begin{corollary} \label{cor:sparse-soft-optim}
Under the observation model and notations in Corollary \ref{cor:sparse-restart} but with a subgassuan parameter $4\sigma^2$ when ${\bs \theta}_{1:n}$ is the re-centered ground truth within a bin discovered by the policy, then with probability atleast $1-2n^{3-\beta/8}$
\begin{align}
     \|\hat{\alpha}_\delta - \alpha \|^2 
     &= \tilde{O}(J\sigma^2).
\end{align}
\end{corollary}

With Corollaries \ref{cor:sparse-restart} and  \ref{cor:sparse-soft-optim}, the proof of Theorem \ref{thm:main} can be readily adapted to give Theorem \ref{thm:exact-sparse}.

\propExactLower*
\begin{proof}
Let $U\{a,b,c\}$ denote a uniform sample from set $\{a,b,c\}$. Consider a ground truth sequence as follows:
\begin{enumerate}
    \item For t=1, ${\bs \theta}_1 = U\{-B,0,B\}$
    \item For t = 2 to $J_n+1$:
    \begin{itemize}
        \item if ${\bs \theta}_{t-1} = -B$, ${\bs \theta}_t = U\{0,B\}$
        \item if ${\bs \theta}_{t-1} = 0$, ${\bs \theta}_t = U\{-B,B\}$
        \item if ${\bs \theta}_{t-1} = B$, ${\bs \theta}_t = U\{-B,0\}$
    \end{itemize}
    \item For $t > J_n + 1$, output ${\bs \theta}_t = {\bs \theta}_{t-1}$
\end{enumerate}

Such a signal will have $\|D^{k+1} {\bs \theta}_{1:n}\|_0 \le J_n$. Let's assume that we reveal this sequence generating process to the learner. Then the Bayes optimal algorithm will suffer a regret of $\Omega(J_n)$.
\end{proof}

\textbf{Extension to higher dimensions}
Let the ground truth $\bs \theta_{1:n}[t] \in \mathbb{R}^d$ and let $ \bs v_i = [\bs \theta_{1:n}[1][i],\ldots,\bs \theta_{1:n}[n][i]], \|D^{k+1}\bs v_i\|_1| \le J_n, \forall i \in [d]$. Then run $d$ instances of \ADAVAW{} where instance $i$ is dedicated to track the sequence $v_i$. By appealing to Theorem \ref{thm:exact-sparse} for each co-ordinate and summing across all $d$ dimensions yields a regret bound of $\tilde O(dJ_n)$.

\section{Adapting to lower orders of k} \label{app:adapt}
Though the theory of offline non parametric regression with squared error loss is well developed for the complete spectrum of function classes $TV^k(C_n)$ with $k \ge 0$, most of the practical interest is often limited to lower orders of $k$ namely $k=0,1,2,3$ (see for eg. \citep{l1tf,tibshirani2014adaptive}). This motivates us to design policies that can perform optimally for these lower TV orders without requiring the knowledge of $k$ beforehand. 

Let $\cE$ be the event that $|\epsilon_t| \le \sigma \sqrt{2 \log (2n^2)}$ for all $t=1,\ldots,n$ where $\epsilon_t$ are as presented in Figure \ref{fig:proto}. By using subgaussian tail inequality and a union bound across all time points, it can be shown that the event $\cE$ happens with probability atleast $1-\frac{1}{n}$.

The basic idea to achieve adaptivity to $k$ is as follows:
\begin{figure}[h!]
	\centering
	\fbox{
		\begin{minipage}{13 cm}
\textbf{Meta-Policy:}
\begin{itemize}
    \item Instantiate \ADAVAW{} for $k=0,1,2,3$ and run them in parallel.
    \item Forecast according to an Exponentially Weighted Averages (EWA) (\citep{BianchiBook2006}) over the predictions made by each of the instances. Set the parameter $\eta$ of EWA to $1/4(B+\sqrt{2 \log (2n^2)})^2$.
\end{itemize}
		\end{minipage}
	}
\end{figure}

We condition on the event $\cE$. The arguments in the proof of Theorem \ref{thm:main} still goes through even if we condition on $\cE$. Let the dynamic regret of \ADAVAW{} for a particular value of $k$ be the random variable $R_n^{(k)}$. The maximum possible value of $R_n^{(k)}$ is $\kappa n$ for some constant $\kappa$.  We have,
\begin{align}
    \E[R_n^{(k)}|\cE]
    &= \int_{-\infty}^{\kappa n} r d\mathbb P(r),\\
    &\le \gamma n^{\frac{1}{2k+3}}C_n^{\frac{2}{2k+3}} + \int_{\gamma n^{\frac{1}{2k+3}}C_n^{\frac{2}{2k+3}}}^{\kappa n} r d\mathbb P(r),\\
    &\le \gamma n^{\frac{1}{2k+3}}C_n^{\frac{2}{2k+3}} + \kappa n \cdot \delta,
\end{align}
for some constant $\gamma$, where last line follows due to Theorem \ref{thm:main}. By choosing $\delta = 1/n$ we get 

\begin{align}
\E[R_n^{(k)}|\cE] 
&= \tilde O \left( n^{\frac{1}{2k+3}}C_n^{\frac{2}{2k+3}} \right). \label{eq:expect}
\end{align}

Let $\hat y_t$, be the output of any forecasting strategy at time $t$. Each expert in the meta-policy suffers a loss $(y_t - \hat y_t)^2$ for appropriate value of $\hat y_t$. Let $\theta_t := \bs \theta_{1:n}[t]$.  we have
\begin{align}
    \sum_{t=1}^{n} \E[(y_t - \hat y_t)^2 | \cE] - \E[(y_t - \theta_t)^2|\cE],
    &=_{(a)} \sum_{t=1}^{n} \E[(\theta_t - \hat y_t)^2 | \cE] - \E[(\hat y_t - \theta_t)^2|\cE] \cE[\epsilon_t | \cE],\\
    &= \sum_{t=1}^{n} \E[(\theta_t - \hat y_t)^2 | \cE], \label{eq:average}
\end{align}
where the last line is simply the expected dynamic regret of the strategy and line (a) is due to independence of $\epsilon_t$ with $\hat y_t$.

 Let the dynamic regret of the meta-policy be denoted as $R_{meta}$. Since squared error loss $(y_t - \hat{y}_t)^2$ is exponentially concave with parameter $1/4(B+\sqrt{2 \log (2n^2)})^2$, Proposition 3.1 of \citep{BianchiBook2006} along with \eqref{eq:expect} and \eqref{eq:average} guarantees that,
\begin{align}
    \E[R_{meta}|\cE]
    &= \log 4 + \tilde{O}\left(\min_{k=0,1,2,3} n^{\frac{1}{2k+3}}\left(n^k \|D^{k+1}\theta_{1:n}\|_1 \right)^{\frac{2}{2k+3}}\right)
\end{align}

Thus we see that expected dynamic regret of the meta-policy adapts to TV order $k$ upto a additive constant of $\log 4$. This additive constant only contributes to a small $O(1/n)$ term if we consider the per round regret.

\section{Problems with padding}
\label{app:pad_problems}
In this section, we explain why some commonly used padding schemes can potentially inflate the $TV^k$ distance of the resulting sequence.

\subsection{Zero padding}
Consider a sequence ${\bs \theta}_{1:t}$ such that best polynomial fit of this sequence is uniformly zero.
Let $\bs \gamma$ be the zero padded version of ${\bs \theta}_{1:t}$ such that length of $\bs \gamma$ is a power of 2. Let $\tilde{\bs \theta} = [{\bs \theta}_{t-k},\ldots,{\bs \theta}_t,0,\ldots,0]^T \in \mathbb{R}^{2k+2}$. We have,
\begin{align}
    (D^{k+1} \bs \gamma)^T
    &= [(D^{k+1}{\bs \theta}_{1:t})^T, (D^{k+1}\tilde{\bs \theta})^T, 0,0,\ldots,0].
\end{align}

Due to \eqref{eqn:pad-bound}, we have $\|{\bs \theta}_{1:t}\|_\infty = O(t^{k}\|D^{k+1} {\bs \theta}_{1:t}\|_1)$. Hence the existence of $\tilde{\bs \theta}$ term makes $\|D^{k+1} \bs \gamma\|_1 = O(t^{k}\|D^{k+1}{\bs \theta}_{1:t}\|_1)$.
\subsection{Mirror image padding}
Let $\bs \gamma$ be the mirror image padded version of the re-centered sequence, ${\bs \theta}_{1:t}$. i.e $\bs \gamma = [\theta_1,\ldots,\theta_t,\theta_t,\theta_{t-1},\ldots]$ such that its length becomes a power of 2. Then,
\begin{align}
    \|D^{k+1}\bs \gamma \|_1
    &= 2\|D^{k+1} {\bs \theta}_{1:t}\|_1 + D^{k+1}[{\bs \theta}_{t-k},\ldots,{\bs \theta}_{t-1},{\bs \theta}_t,{\bs \theta}_t,{\bs \theta}_{t-1},\ldots,{\bs \theta}_{t-k}]^T,\\
    &= 2\|D^{k+1} {\bs \theta}_{1:t}\|_1 + O(t^k \|D^{k+1}{\bs \theta}_{1:t}\|_1),
\end{align}
where the last line follows from \eqref{eqn:pad-bound}.

\section{Technical Lemmas}
\begin{lemma}
\label{lem:polydet}
The procedure \texttt{CalcDetRecurse} in \citep{dingle} is sound.
\end{lemma}
\begin{proof}
We use induction on the dimension of the input square matrix. 

\textbf{Base case:} when $d = 3$. Assume that $e[0][0]$ is non-zero. Let the matrix be given by
\begin{gather}
    \bs{X}
    =
    \begin{bmatrix}
    e_{00} & e_{01} & e_{02}\\
    e_{10} & e_{11} & e_{12}\\
    e_{20} & e_{21} & e_{22}
    \end{bmatrix}
\end{gather}

The idea is to convert $\bs{X}$ to an upper triangular matrix. Define:
\begin{gather}
    \bs{Y}
    =
    \begin{bmatrix}
    1 & \frac{e_{01}}{e_{00}} & \frac{e_{02}}{e_{00}}\\
    e_{10} & e_{11} & e_{12}\\
    e_{20} & e_{21} & e_{22}
    \end{bmatrix}
\end{gather}

So that $\det(\bs{Y}) = \frac{\det(\bs{X})}{e_{00}}$. Applying elementary row operations we get
\begin{gather}
    \det(\bs{Y})
    =
    \begin{vmatrix}
    1 & \frac{e_{01}}{e_{00}} & \frac{e_{02}}{e_{00}}\\
    0 & e_{11}-e_{10}\frac{e_{01}}{e_{00}} & e_{12} - e_{10}\frac{e_{01}}{e_{00}}\\
    0 & e_{21}-e_{20}\frac{e_{01}}{e_{00}} & e_{22} - -e_{20}\frac{e_{01}}{e_{00}}
    \end{vmatrix}
\end{gather}

The inner loop in the procedure \texttt{CalcDetRecurse} computes the determinant of the inner $2 \times 2$ sub-matrix by considering the numerator of the fractional terms. Hence the value $v$ return by the recursive call is $\det(\bs{Y}[1:][1:]) e_{00}^2$. So $\det(X) = e_{00} \frac{v}{e_{00}^2} = \frac{v}{e_{00}}$. This is precisely the value returned by the procedure after the final division loop.

When $e_{00}$ is zero, we can swap it with the row whose first element is non-zero and apply the arguments above. If such a swap is not possible, the procedure correctly recognizes the determinant as zero.

\textbf{Inductive case:} Assume that procedure is sound for matrices upto dimension $n$. Now define $\bs{Y}$ as before to set the element $e_{00}$ to one. By similar arguments we obtain that value $v$ returned by the recursive call is $\det(\bs{Y}[1:][1:]) e_{00}^n$. Thus we obtain $\det(\bs{X}) = \frac{v}{e_{00}^{n-1}}$. This division is performed at the final loop of the procedure.

Here also when $e_{00}$ is zero, the swapping argument similar to the base case can be applied.

\end{proof}

Consider OLS fit on the inputs $(\boldsymbol{x_1},y_1),\ldots,\boldsymbol{x_t},y_t)$ where the features $\boldsymbol{x_j} = [1,j,\ldots,j^m]^T$ and the responses obey $\max_{i=1,\ldots,t}|y_i| = B$. Let the design matrix be 
\begin{align}
    \boldsymbol{X}_t = [\boldsymbol{x}_1,\ldots,\boldsymbol{x}_t]^T.
\end{align}

\begin{lemma}
\label{lem:det_degree}
$\det(\bs{X_t}^T\bs{X_t})$ is a polynomial in $t$ with degree atmost $(k+1)^2$.
\end{lemma}
\begin{proof}
The procedure \texttt{CalcDegreeOfDet} in \citep{dingle} can be used to upperbound the degree of determinant. It assumes that while doing the subtractions in procedure \texttt{CalcDetRecurse}, the highest degree terms in the corresponding polynomials do not cancel out. 

Let $m=k+1$. Observe that $\bs{X}_t^T\bs{X_t}$ can be compactly written as
\begin{gather}
    \label{eqn:matsum}
    \bs{X}_t^T\bs{X_t}
    =
    \begin{bmatrix}
    S_0(t) & S_1(t) &\dots & S_{m-1}(t)\\
    \vdots &\vdots &\ddots &\vdots\\
    S_{m-1}(t) & S_m(t) &\dots &S_{2m-2}(t)
    \end{bmatrix},
\end{gather}
where $S_{p}(t) = \sum_{n=1}^{t} n^p$.

Let's run procedure \texttt{CalcDegreeOfDet} on an $m \times m$ matrix $\bs{D}$ of degrees arising from $\bs{X}_t^T\bs{X}_t$ as below.
\begin{gather}
    \bs{D}
    =
    \begin{bmatrix}
    1 & 2 &\dots &m\\
    \vdots &\vdots &\ddots &\vdots\\
    m & m+1 &\dots &2m-1
    \end{bmatrix}
\end{gather}

Let's define a seed sequence $\{s\}_i$ as the sequence of numbers that can be found the main diagonal of a given matrix, excluding the element at the bottom right corner. The seed sequnece of $\bs{D}$ is simply $1,3,\ldots,2m-3$. Let $T_i$ be the element at index $(0,0)$ for the matrix in the $i^{th}$ recursive call. Note that $T_1 = 1$. Tracing the steps through the recursion we get
\begin{align}
    T_2 &= s_2 + T_1\\
    T_3 &= s_2 + T_2 + T_1\\
    &\vdots \notag \\
    T_{m-1} &= s_{k-1} + T_{k-2} + \ldots + T_1 
\end{align}

In $m-1$ calls, we will be left with a $2 \times 2$ matrix whose entries are
\begin{gather}
    \begin{bmatrix}
    T_{m-1} & 1 + T_{m-1}\\
    1 + T_{m-1} & 2 + T_{m-1}
    \end{bmatrix}
\end{gather}

Now let's start with the winding up procedure. There are $k-3$ wind-ups that need to be performed. Let $u_t$ be the wound up value from the $t^{th}$ winding up step. We have,
\begin{align}
    u_{m-2} &= 2 + 2T_{m-1} - T_{m-2}\\
    u_{m-3} &= u_{m-2} - 2T_{m-3}\\
    u_{m-4} &= u_{m-3} - 3T_{m-4}\\
    &\vdots \notag \\
    u_1 &= u_2 - (m-2)T_1
\end{align}

Note that $u_1$ is the final output produced by the topmost call to \texttt{CalcDegreeOfDet} procedure. These systems can be unrolled to get
\begin{align}
    u_1 
    &= 2 + 2T_{m-1} - (T_{m-2}+ 2T_{m-3} + \ldots + (m-2)T_1\\
    &= 2 + s_{m-1} + \sum_{i=1}^{m-1} s_i \label{eqn:f1}
\end{align}

Now using explicit expressions for seed sequence $\{s\}_i$ we get
\begin{align}
    u_1 
    &= 2 + 2m-3 + (m-1)^2\\
    &= m^2\\
    &= (k+1)^2
\end{align}
\end{proof}

\begin{lemma}
\label{lem:faul}
Let $S_p(t)$ be a polynomial in $t$ defined as $S_p(t) = \sum_{n=1}^{t} n^p$ where $p$ is a non-negative integer. Then,
\begin{align}
    (-1)^{p-1} S_p(t-1) = S_p(-t)
\end{align}
\end{lemma}
\begin{proof}
For $a(t) = \frac{t(t+1)}{2}$, Faulhaber's formula states that
\begin{align}
    \sum_{n=1}^{t} n^p = \sum_{i=1}^{(p-1)/2} c_i a(t)^{(p+1)/2},
\end{align}
when $p$ is odd and
\begin{align}
    \sum_{n=1}^{t} n^p = \frac{t+0.5}{p+1}\sum_{i=1}^{p/2} (i+1) c_i a(t)^{p/2},
\end{align}
when $p$ is even. The the explicit form of $c_i$ can be expressed in terms of Bernoulli numbers. 

Note that $a(-t) = a(t-1)$. Substituting this in the formulas yields the lemma.
\end{proof}

\begin{lemma}
\label{lem:det_poly}
For a universal constant $H(m)$ that depends only on $m=k+1$,
\begin{align}
    \det(\bs{X}_t^T\bs{X}_t)
    &=
    H(m)\: t^m \prod_{i=2}^{m} \left(t^2 - (i-1)^2\right)^{m-i+1}
\end{align}
\end{lemma}
\begin{proof}
The strategy is to characterize the roots of determinant. For brevity let's denote $\bs{Z}_t = \bs{X}_t^T\bs{X}_t$. Observe that
\begin{align}
    \bs{Z}_t
    &= \sum_{i=1}^{t} \bs{x}_i\bs{x}_i^T \label{eqn:pot},
\end{align}
where $x_i = [1,\ldots,i^{m-1}]$. Each update $\bs{x}_i\bs{x}_i^T$ increases the rank by atmost 1. After $m$ such updates $\bs{X}_m$ becomes a square Vandermonde matrix formed by the sequence $\{1,2,\ldots,m\}$. Since all of the elements in the sequence are distinct $\bs{X}_m$ is full rank and so is $\bs{Z}_m$. This implies that each such update $\bs{x}_i\bs{x}_i^T$ for $i \le m$ increased the rank by exactly one.

We can view the equation \eqref{eqn:pot} as a quantity that evolves in time. For $1 \le i \le m-1$, there exists $m-i$ rows in $\bs{Z}_i$ that are linearly dependent. This means $t=i$ is a root of $\det(\bs{Z}_t)$ with multiplicity $(m-i)$. By defining $\bs{x}_0 = [0,\ldots,0]^T$  for the initial case $t=0$, all the rows are simply zeroes and multiplicity of the root $t=0$ is $m$. Thus we have established that $t^m \prod_{i=2}^{m} \left(t - (i-1)\right)^{m-i+1}$ is a sub-expression of $\det(\bs{Z}_t)$.

Let's view $\bs{Z}_t$ as a function of $t$ with $t\in \mathbb{R}$ as displayed in \eqref{eqn:matsum}. Put $t = -t'$ in \eqref{eqn:matsum}. Then we have,

\begin{gather}
    \bs{Z}(t')
    =
    \begin{bmatrix}
    S_0(-t') & S_1(-t') &\dots & S_{m-1}(-t')\\
    \vdots &\vdots &\ddots &\vdots\\
    S_{m-1}(-t') & S_m(-t') &\dots &S_{2m-2}(-t')
    \end{bmatrix}.
\end{gather}

Hence showing $t'=a$ is a root of $\bs{Z}(t')$ implies that $t=-a$ is a root of $\bs{Z}_t$.
We have
\begin{gather}
    \det(\bs{Z}(t'))
    = (-1)^m
    \begin{vmatrix}
    -S_0(-t') & -S_1(-t') &\dots & -S_{m-1}(-t')\\
    \vdots &\vdots &\ddots &\vdots\\
    -S_{m-1}(-t') & -S_m(-t') &\dots &-S_{2m-2}(-t')
    \end{vmatrix}
\end{gather}

Consider
\begin{gather}
    \det(\bs{\tilde Z}(t'))
    = 
    \begin{vmatrix}
    -S_0(-t') & -S_1(-t') &\dots & -S_{m-1}(-t')\\
    \vdots &\vdots &\ddots &\vdots\\
    -S_{m-1}(-t') & -S_m(-t') &\dots &-S_{2m-2}(-t')
    \end{vmatrix}
\end{gather}

When $t'$ is a non-negative integer, lemma \ref{lem:faul} implies that the elements in the matrix above are result of the summation:
\begin{align}
    \sum_{i=0}^{t'-1} (-i)^p
    &= (-1)^p S_p(t'-1)\\
    &= -S_p(-t'),
\end{align}

where we adopt the convention $0^0 = 1$.

Thus we have,
\begin{align}
    \bs{\tilde Z}(t')
    &= \sum_{i=1}^{t'} \bs{x'}_i\bs{x'}_i^T,
\end{align}
where $\bs{x'}_i = [1,-(i-1),\ldots,\left(-(i-1)\right)^{m-1}]$. Let $\bs{X'}_t = [\boldsymbol{x'}_1,\ldots,\boldsymbol{x'}_t]^T.$

After $m$ updates, we have that $\bs{X'}_m$ is a square Vandermonde matrix defined by the sequence $\{0,-1,\ldots,-(m-1)\}$. Since each of the elements are distinct, this a full rank matrix and so each update $\bs{x'}_i\bs{x'}_i^T$ for $i \le m$ increased the rank by exactly one leading to $\bs{\tilde Z}(m)$ being full rank.

Using similar arguments as above we see that $t'=i$ is a root of $\det(\bs{\tilde Z}(t'))$ with multiplicity $(m-i)$. This in turn imply that $t=-i$ is a root of $\det(\bs{Z}_t)$ with multiplicity $(m-i)$. Now we have established that $t^m \prod_{i=2}^{m} \left(t^2 - (i-1)^2\right)^{m-i+1}$ is a sub-expression of $\det(\bs{Z}_t)$. By lemma \ref{lem:det_degree} we conclude that we have found all roots of the determinant and no further terms depending $t$ can be there.

\end{proof}

\begin{remark}
We conjecture that the universal constant $H(m)$ in lemma \ref{lem:det_poly} is the determinant of Hilbert matrix of order $m$.
\end{remark}

\begin{definition}
Let $\bs{H}(t)$ be a square matrix with each entry $r_{ij}(t) = \frac{n_{ij}(t)}{d_{ij}(t)}$ for polynomials $n_{ij}(t)$ and $d_{ij}(t)$. We say $r_{ij}(t)$ is \emph{Hilbert-like} if $r_{ij}(t) = O\left(\frac{1}{t^{i+j-1}}\right)$ for all $i,j$.
\end{definition}

\begin{lemma}
\label{lem:hilbert}
All the elements of $\left(\bs{X}_t^T\bs{X}_t\right)^{-1}$ are Hilbert-like when $t \ge m=k+1$.
\end{lemma}
\begin{proof}
Computation of inverse is essentially a computation of determinants of the matrix and its minors. Each element $(i,j)$ of an inverse matrix is a rational function with numerator being determinant of minor $M_{ij}$ and denominator being the determinant of the original symmetric matrix. 

Let $\bs{Z}_t = \bs{X}_t^T\bs{X}_t$ When $t \ge m$ we have from lemma \ref{lem:det_poly} that $\det(\bs{Z}_t) = \Omega(t^{m^2})$. So it is sufficient to show that $\det(M_{ij}$ is $O(t^{m^2+1-i-j})$. The strategy we follow is same of that in lemma \ref{lem:det_degree}. 

We follow a 1 based indexing. Since $\bs{Z}_t$ is symmetric, it is enough to compute the minors when $1 \le i \le j \le m$.

\textbf{case 1: } Consider $\det(M_{ij})$ when $1 < i < j < m-1$. Following the same notations as in the prood of lemma \ref{lem:det_poly}, after $m-2$ calls to \texttt{CalDegreeOfDet} we end up with a matrix below.

\begin{gather}
    \label{eqn:fin1}
    \bs{F} 
    = 
    \begin{bmatrix}
    T_{m-2} & 1 + T_{m-2}\\
    1 + T_{m-2} & 2 + T_{m-2}
    \end{bmatrix}
\end{gather}

The corresponding seed sequence $\{s\}_i$ is $\{1,3,5,\ldots,2i-3,2i,2i+2,\ldots,2j-2,2j+1,2j+3,\ldots,2m-3\}$. The jumps in the progression is attributed to the deletion of row $i$ and column $j$ for obtaining minor $M_{ij}$.

The final output $u_1$, from the topmost call to \texttt{CalDegreeOfDet} is then given by
\begin{align}
    u_1 
    &= s_{m-2} + \sum_{i=1}^{m-2} s_i\\
    &= 2 + (2m-3) + (i-1)^2 + (j-i)(j+i-1) + (m+j-1)(m-j-1),\\
    &= m^2 + 1 - i - j.
\end{align}
So $\det(M_{ij})$ is $O(t^{m^2+1-i-j})$ where the constant in the big-oh only dependents on $m$.

\textbf{case 2: } ($1< i < j = m-1$). After $m-2$ recursion calls we get the matrix below.

\begin{gather}
    \label{eqn:fin2}
    \bs{F} 
    = 
    \begin{bmatrix}
    T_{m-2} & 2 + T_{m-2}\\
    1 + T_{m-2} & 3 + T_{m-2}
    \end{bmatrix}
\end{gather}

The seed sequence $\{s\}_i$ is $\{1,3,\ldots,2i-3,2i,\ldots,2j-2 \}$. So

\begin{align}
    u_1
    &= 3 + s_{m-2} + \sum_{i=1}^{m-2} s_i, \\
    &= 3 + (2m-4) + (i-1)^2 + (j-i)(j+i-1),\\
    &=  m^2 + 1 - i - j.
\end{align}

So $\det(M_{ij})$ is $O(t^{m^2+1-i-j})$.

\textbf{case 3: } ($1 < i = j < k-1$).

The seed sequence $\{s\}_i$ is $\{1,3,\ldots,2i-3, 2i+1,\ldots,2m-3 \}$. At the last step we get a matrix as in equation \eqref{eqn:fin1}. Hence,

\begin{align}
    u_1 
    &= 2 + s_{m-2} + \sum_{i=1}^{m-2} s_i,\\
    &= 2 + (2m-3) + (i-1)^2 + (m-i-1)(2i+1+m-i-2),\\
    &= m^2 + 1 - i - j.
\end{align}

So $\det(M_{ij})$ is $O(t^{m^2+1-i-j})$.

\textbf{case 4: } ($i=j=m-1$).

The seed sequence $\{s\}_i$ is $\{1,3,\ldots,2i-3\}$. At the last step we get a matrix below.

\begin{gather}
    \label{eqn:fin3}
    \bs{F} 
    = 
    \begin{bmatrix}
    T_{m-2} & 2 + T_{m-2}\\
    2 + T_{m-2} & 3 + T_{m-2}
    \end{bmatrix}
\end{gather}

So,
\begin{align}
    u_1 
    &= 4 + s_{m-2} + \sum_{i=1}^{m-2} s_i,\\
    &= 2 + (2i-3) + (i-1)^2,\\
    &= m^2 + 1 - i - j.    
\end{align}

So $\det(M_{ij})$ is $O(t^{m^2+1-i-j})$.

\textbf{case 5: } ($i=j=m$).

The seed sequence $\{s\}_i$ is $\{1,3,\ldots,2m-5 \}$. At the last step we get a matrix as in equation \eqref{eqn:fin1}.

\begin{align}
    u_1 
    &= 2 + s_{m-2} + \sum_{i=1}^{m-2} s_i,\\
    &= 2 + (2m-5) + (m-2)^2,\\
    &= m^2 + 1 - i - j.
\end{align}

So $\det(M_{ij})$ is $O(t^{m^2+1-i-j})$.

\textbf{case 6: } ($i=j=1$).

The seed sequence $\{s\}_i$ is $\{3,\ldots,2m-3 \}$. At the last step we get a matrix as in equation \eqref{eqn:fin1}.

\begin{align}
    u_1 
    &= 2 + s_{m-2} + \sum_{i=1}^{m-2} s_i,\\
    &= 2 + (2m-3) + (m-1)^2 - 1,\\
    &= m^2 + 1 - i - j.
\end{align}

So $\det(M_{ij})$ is $O(t^{m^2+1-i-j})$.

\textbf{case 7: } ($1<i<k-1<j=k$).

The seed sequence $\{s\}_i$ is $\{1,\ldots,2i-3,2i,\ldots,2m-4 \}$. At the last step we get a matrix as in equation \eqref{eqn:fin1}.

So,

\begin{align}
    u_1 
    &= 2 + s_{m-2} + \sum_{i=1}^{m-2} s_i,\\
    &= 2 + (2m-4) + (i-1)^2 + (m-i-1) (2i + k - i - 2),\\
    &= m^2 + 1 - i - j.
\end{align}

So $\det(M_{ij})$ is $O(t^{m^2+1-i-j})$.

\textbf{case 8: } ($i=1, j=m$).

The seed sequence $\{s\}_i$ is $\{2,\ldots,2m-4\}$. At the last step we get a matrix as in equation \eqref{eqn:fin1}.

So,

\begin{align}
    u_1 
    &= 2 + s_{m-2} + \sum_{i=1}^{m-2} s_i,\\
    &= 2 + (2m-4) + (m-2)(2+m-3),\\
    &= m^2 + 1 - i - j.
\end{align}

So $\det(M_{ij})$ is $O(t^{m^2+1-i-j})$.

\textbf{case 9: } ($i=1, j=m-1$).

The seed sequence $\{s\}_i$ is $\{2,\ldots,2m-4\}$. At the last step we get a matrix as in equation \eqref{eqn:fin2}.

So,

\begin{align}
    u_1 
    &= 3 + s_{m-2} + \sum_{i=1}^{m-2} s_i,\\
    &= 3 + (2m-4) + (m-2)(2+m-3),\\
    &= m^2 + 1 - i - j.
\end{align}

So $\det(M_{ij})$ is $O(t^{m^2+1-i-j})$.

\textbf{case 10: } ($i=1 < j < m-1$).

The seed sequence $\{s\}_i$ is $\{2,\ldots,2j-2,2j+1,\ldots,2m-3\}$. At the last step we get a matrix as in equation \eqref{eqn:fin1}.

So,

\begin{align}
    u_1 
    &= 2 + s_{m-2} + \sum_{i=1}^{m-2} s_i,\\
    &= 2 + (2m-3) + (j-1)(2+j-2) + (m-j-1)(2j+1+(m-j-2)),\\
    &= m^2 + 1 - i - j.
\end{align}

So $\det(M_{ij})$ is $O(t^{m^2+1-i-j})$.

\end{proof}

With the above lemma, the following Corollary can be readily verified.
\begin{corollary} \label{cor:bounded_OLScoeffs}
When $\bs \theta_{1:n}$ is such that $\| \bs \theta_{1:n} \|_\infty \le B = O(1)$, we have $\|\left(\bs{X}_t^T\bs{X}_t\right)^{-1}\bs{X}_t^T  \bs \theta_{1:n}\|_2 = O(1)$.
\end{corollary}

\end{document}